\documentclass[12pt, a4paper, oneside]{article}

\usepackage[utf8]{inputenc}
\usepackage[T1]{fontenc}
\usepackage{times}
\usepackage{graphicx}
\usepackage{amsmath, amssymb, amsthm}
\usepackage{booktabs}
\usepackage{tabularx}
\usepackage{longtable}
\usepackage[hypertexnames=false]{hyperref}
\usepackage{cite}
\usepackage{url}
\usepackage{etoolbox}
\usepackage{setspace}
\usepackage[left=1in, right=1in, top=1in, bottom=1in]{geometry}
\usepackage{listings}
\usepackage{xcolor}
\usepackage{algorithm}
\usepackage[noend]{algpseudocode}

\hypersetup{
    pdftitle={Architectural Design Decisions in AI Agent Harnesses: An Empirical Study of 70 Publicly Available Projects},
    pdfauthor={Anonymous},
    pdfsubject={AI Agent Framework Design Space Analysis},
    pdfkeywords={AI Agents, Agent Frameworks, Software Architecture, Design Space, Empirical Study}
}

\title{Architectural Design Decisions in AI Agent Harnesses}

\author{
    Hu Wei$^{1}$\\
    {1990huwei@sina.com}$^{1}$
}
\date{}

\begin{document}

\maketitle

\begin{abstract}
AI agent systems increasingly rely on reusable non-LLM engineering infrastructure that packages tool mediation, context handling, delegation, safety control, and orchestration. Yet the architectural design decisions in this surrounding infrastructure remain understudied. This paper presents a protocol-guided, source-grounded empirical study of 70 publicly available agent-system projects, addressing three questions: which design-decision dimensions recur across projects, which co-occurrences structure those decisions, and which typical architectural patterns emerge. Methodologically, we contribute a transparent investigation procedure for analyzing heterogeneous agent-system corpora through source-code and technical-material reading. Empirically, we identify five recurring design dimensions (subagent architecture, context management, tool systems, safety mechanisms, and orchestration) and find that the corpus favors file-persistent, hybrid, and hierarchical context strategies; registry-oriented tool systems remain dominant while MCP- and plugin-oriented extensions are emerging; and intermediate isolation is common but high-assurance audit is rare. Cross-project co-occurrence analysis reveals that deeper coordination pairs with more explicit context services, stronger execution environments with more structured governance, and formalized tool-registration boundaries with broader ecosystem ambitions. We synthesize five recurring architectural patterns spanning lightweight tools, balanced CLI frameworks, multi-agent orchestrators, enterprise systems, and scenario-verticalized projects. The result provides an evidence-based account of architectural regularities in agent-system engineering, with grounded guidance for framework designers, selectors, and researchers.
\end{abstract}



\section{Introduction}

\subsection{Motivation}
Artificial intelligence agents built on large language models (LLMs) are rapidly becoming a practical paradigm for automated task execution. As these systems move from demonstrations to persistent software, they depend on substantial non-LLM engineering infrastructure: tool invocation, context retention, execution control, delegation, workspace access, and coordination across multiple components. Across the ecosystem, that infrastructure appears in projects described as agent systems, agent frameworks, agent platforms, CLIs, runtimes, and related forms.

In recent years, the public ecosystem has produced a large and diverse set of such projects. Projects such as OpenHands~\cite{openhands}, OpenClaw~\cite{openclaw}, Claude Code CLI~\cite{claude_code}, docker-agent~\cite{docker_agent}, and AgentPool illustrate markedly different architectural approaches to operationalizing Agent systems. Some remain single-agent and tightly scoped, whereas others support explicit subagent hierarchies. Some rely on simple session-bound context handling, whereas others implement layered persistence and retrieval. Some expose broad execution power with minimal controls, whereas others integrate sandboxing, approval, and auditing into the surrounding engineering layer. This diversity is not incidental background variation. It shows that the ecosystem is already exploring multiple competing answers to the core infrastructural question of how Agent behavior should be made reusable, governable, and extensible.

In this paper, we use the term ``Agent harness'' only as a loose working label for this non-LLM engineering layer around Agent behavior. It is not intended as a strict definition or a claim to introduce a new formal category. The projects in our corpus variously present themselves as agent systems, frameworks, platforms, CLIs, runtimes, or products. ``Harness'' is therefore used for analytical convenience: it lets us refer compactly to the surrounding infrastructure for tool mediation, state and context handling, delegation structure, execution boundaries, and governance controls without claiming that all projects belong to one explicitly named software class.

Despite this growth, there is still no systematic account of the architectural design decisions that characterize this surrounding infrastructure across projects. Prior work has examined agent algorithms, including reasoning strategies~\cite{react}, generative social behavior~\cite{generative_agents}, and tool-use mechanisms~\cite{toolformer}, while many individual frameworks have been proposed and released~\cite{langchain,autogpt,crewai}. However, this literature offers limited leverage for answering infrastructure-level questions such as which decisions recur across projects, which decisions tend to be bundled together, and how different product positions relate to different architectural commitments. As a result, the field knows far more about what agents can do than about how their surrounding engineering systems are architecturally assembled.

This gap creates both practical and research challenges. Framework developers lack empirical guidance when making architectural decisions and may therefore revisit tradeoffs that have already been explored elsewhere. Practitioners selecting frameworks have limited support for understanding the implications of different architectural bundles. Researchers, meanwhile, lack a shared vocabulary for describing and comparing the architectural organization of these non-LLM engineering layers, which in turn makes it difficult to test or reject common simplifications such as ``language determines architecture'' or ``capability growth naturally brings governance maturity.'' Without a comparative empirical account, architectural variation remains visible but analytically underorganized.

This paper addresses this gap through a protocol-guided, source-grounded empirical study of 70 publicly available agent-system projects. By examining recurring design decisions across key architectural dimensions, we identify recurrent architectural regularities, document observed co-occurrences and analytically informative non-co-occurrences among those decisions, characterize patterns that help explain how projects with different complexity profiles and product positioning assemble their architectures, and formulate analytically useful conjectures that later work can test against expanded or longitudinal corpora. The broader claim is that the non-LLM engineering layer around Agent systems has become sufficiently consequential to warrant direct architectural study rather than incidental treatment as implementation detail.

\subsection{Research Questions}
We formulate three research questions that guide our empirical analysis:

\begin{itemize}
    \item \textbf{RQ1 (Design Space):} What are the key design-decision dimensions in Agent harness frameworks, and what option spectra recur within each dimension?

    \item \textbf{RQ2 (Co-occurrences):} What recurring co-occurrences and non-co-occurrences can be observed between design decisions across frameworks?

    \item \textbf{RQ3 (Selection and Positioning):} How do projects with different complexity levels and product positioning choose within the design space, and what typical patterns help explain these decisions?
\end{itemize}

Taken together, these questions move from empirical identification to cross-project relationship analysis to practical guidance. This progression reflects our goal of explaining recurrent architectural structure rather than merely cataloging frameworks or ranking them by overall quality.

\subsection{Contributions}
This paper makes four contributions:

\begin{enumerate}
    \item \textbf{Empirical Characterization Grounded in a Transparent Investigation Procedure:} We identify the focal design decisions that recur most consistently across publicly available Agent execution infrastructures, showing that subagent architecture, context management, tool systems, safety mechanisms, and orchestration provide a coherent empirical basis for analyzing this software category, and we ground that characterization in a transparent protocol-guided investigation procedure with explicit evidence trails and comparable coding logic.

    \item \textbf{Observed Co-occurrences and Non-Co-occurrences:} We report recurring co-occurrences and analytically informative non-co-occurrences among those decisions, showing that the design space is structured by recognizable decision bundles rather than by isolated feature choices and that several simplifying assumptions, including ``language determines architecture,'' ``use case fixes complexity,'' and ``capability growth implies governance maturity,'' do not survive cross-project comparison.

    \item \textbf{Pattern Synthesis for Comparison and Selection:} We synthesize typical architectural patterns that explain how projects with different complexity envelopes, governance commitments, and product positioning assemble different combinations of decisions, thereby turning cross-project variation into a reusable comparative framework for explanation, development, and selection.

    \item \textbf{Auditable Comparative Research Package:} We provide a documented corpus boundary, search and screening protocol, and operational codebook in the appendices so that later studies can audit, revisit, refine, or extend the present comparison without relying on an opaque or purely impressionistic research process.
\end{enumerate}

\subsection{Paper Structure}
The remainder of this paper is organized as follows. Section~\ref{sec:relatedwork} reviews related work on AI Agent systems, framework infrastructure, and software architecture design-space analysis. Section~\ref{sec:methodology} describes our corpus, coding dimensions, and extraction procedure. Section~\ref{sec:designspace} addresses RQ1 by describing the key design-decision dimensions and their option spectra. Section~\ref{sec:correlations} addresses RQ2 by presenting both recurring co-occurrences and important non-co-occurrences between design decisions. Section~\ref{sec:patterns} addresses RQ3 by summarizing typical combinations of decisions across projects with different complexity and positioning. Section~\ref{sec:discussion} relates the findings to prior work, practical decision making, and validity considerations. Section~\ref{sec:conclusion} summarizes the study and outlines future research directions.


\section{Related Work}
\label{sec:relatedwork}

\subsection{AI Agent Systems}
The concept of AI agents has evolved substantially from early rule-based systems to contemporary LLM-powered autonomous agents. A major milestone was the introduction of the ReAct framework~\cite{react}, which demonstrated that tightly coupling reasoning and action can improve performance across diverse tasks. This work helped establish the paradigm of agents that explicitly reason about their actions while executing them, thereby defining one of the control pressures that later execution infrastructures must support in practice. At the same time, ReAct did not address how such reasoning-action loops are packaged into reusable systems with persistent state, execution boundaries, and tool mediation across projects.

Park et al. introduced Generative Agents~\cite{generative_agents}, demonstrating that simulated characters powered by LLMs could exhibit emergent social behaviors in sandboxed environments. This work highlighted the importance of long-term memory and interaction continuity in agent systems, but it did not ask how those requirements are operationalized across reusable software substrates. In architectural terms, it made memory a central capability pressure without showing how different projects engineer persistence, summarization, and retrieval around that pressure.

Schick et al. proposed Toolformer~\cite{toolformer}, showing that language models can learn to use external tools through self-supervised learning. This work established tool use as a core capability of LLM-based agents and influenced subsequent framework design, yet it still leaves open the architectural question of how tool mediation is packaged into reusable infrastructures. The capability result therefore intensifies, rather than resolves, the comparative question of how projects define, register, discover, and safely execute tools in practice.

Multi-agent systems have likewise been studied extensively. Qian et al. explored communicative agents for software development~\cite{chatdev}, demonstrating that multi-agent collaboration can address tasks that exceed the practical scope of single-agent systems. Again, the capability result is important, but it does not by itself explain how reusable frameworks encode delegation boundaries, coordination roles, or state transfer. The existence of multi-agent capability thus creates a coordination burden whose implementation still has to be compared at the level of surrounding engineering systems.

Recent surveys have characterized the architecture of LLM-based agents more broadly. Wang et al. provided a comprehensive survey of LLM-based agents~\cite{agent_survey}, identifying key components including planning, memory, tool use, and perception. Hao et al. surveyed planning and reasoning approaches in LLMs~\cite{plan_act}, categorizing different strategies agents employ for task decomposition and execution. These studies are valuable because they identify the functional burdens an Agent system may need to absorb. However, they still operate primarily at the level of component functions or agent capabilities rather than at the level of cross-project comparison of how those burdens are assembled into reusable project architectures.

These prior works focus primarily on agent algorithms and behaviors. In contrast, our study examines the surrounding non-LLM engineering layer that operationalizes these capabilities, namely the infrastructure for execution environments, memory management, tool integration, governance boundaries, and reusable coordination structure. The key gap is therefore not the absence of Agent research, but the absence of a comparative architectural account of how these infrastructural commitments are assembled across projects.

\subsection{Agent Frameworks and Platforms}
A diverse ecosystem of Agent frameworks and platforms has emerged to support the development of LLM-powered applications. For analytical purposes, this ecosystem can be read through two contrasting emphases: \textbf{application frameworks}, which primarily provide programming abstractions for building agent applications, and \textbf{surrounding execution infrastructure}, which more explicitly organizes the reusable runtime conditions within which agents operate. In this paper, ``Agent harness'' is used only as a loose shorthand for this latter emphasis rather than as a strict category definition.

Application frameworks include LangChain~\cite{langchain} and its successor LangGraph~\cite{langgraph}, which expose chains, graphs, agents, and memory abstractions to application developers. LlamaIndex~\cite{llamaindex} similarly foregrounds data-centric retrieval, memory, and agent composition over user data. AutoGPT~\cite{autogpt} demonstrated autonomous goal-directed behavior through iterative prompting and tool use. CrewAI~\cite{crewai} introduced role-based multi-agent collaboration, MetaGPT~\cite{metagpt} proposed structured coordination through generated artifacts, and AgentScope~\cite{agentscope} emphasized agent-oriented programming for multi-agent applications. These systems are important because they popularized Agent development patterns, but they are not primarily comparative studies of the infrastructure choices that make Agent operation reusable, governable, and deployable.

Projects in this second group foreground the infrastructure layer more explicitly. They decide how tools are registered and mediated, how execution is isolated, how context and state are retained, how delegation is organized, and how safety checks are enforced. OpenHands~\cite{openhands} emphasizes repository integration and execution support for software-development agents. SWE-agent~\cite{swe_agent} provides an agent-computer interface specialized for code-repository problem solving. OpenClaw~\cite{openclaw} offers a TypeScript-based execution substrate with multiple architectural styles. Claude Code CLI~\cite{claude_code} provides a coding-focused command-line harness with sandboxed execution. The docker-agent project~\cite{docker_agent} explores container-based execution as an explicit basis for operational reliability. What is still missing, however, is a comparative account of which infrastructural decisions recur across such systems and how those decisions combine into recognizable architectural bundles.

Benchmarking efforts have likewise focused on Agent capability rather than harness architecture. AgentBench~\cite{agentbench} evaluates LLMs as agents across multiple domains. SWE-Bench~\cite{swe_bench} focuses on software engineering tasks. WebArena~\cite{webarena} provides a realistic web environment for agent evaluation, and more recent work such as Tau-Bench~\cite{tau_bench} expands evaluation toward planning and memory-intensive behavior. These benchmarks are indispensable for assessing what agents can do, but they offer limited leverage for explaining why two frameworks with similar capability goals may adopt very different execution, memory, delegation, or governance structures. In other words, benchmark success can reveal behavioral performance while still leaving execution-substrate decisions undertheorized.

Recent safety-oriented work points in a similar direction. Studies of MCP security~\cite{mcp_safety,mcp_empirical} and broader surveys of LLM-agent risks~\cite{gan_agent_risks} show that tool mediation, protocol boundaries, and operational safeguards matter greatly for downstream reliability. Yet these studies still do not ask the comparative architectural question at the center of this paper: which execution-substrate decisions recur across frameworks, which combinations are common, and which expected relationships fail to recur.

The resulting gap is therefore not simply a missing list of frameworks. It is a missing comparative account of the infrastructure decisions through which Agent systems are operationalized. That gap motivates the present study.

\subsection{Software Architecture Design Space Analysis}
Design space analysis is a well-established methodology in software engineering for studying architectural alternatives and recurring decision patterns. Rather than treating systems as isolated artifacts, it examines the set of design decisions that recur within a domain, the relationships among those decisions, and the tradeoffs that follow from them.

This perspective aligns with software-architecture research that treats architectural decisions as first-class analytical objects. Jansen and Bosch argued that software architecture can be understood as a set of explicit architectural design decisions rather than only as a static arrangement of components and connectors~\cite{jansen_bosch_decisions}. Later work on architectural-decision knowledge showed that documenting, comparing, and revisiting such decisions is itself a mature research concern rather than an incidental by-product of implementation~\cite{tofan_decision_mapping}.

Hogan et al. conducted a comprehensive survey of software architecture design-space exploration, categorizing methods for systematically studying design alternatives. Their work provides an important methodological foundation for our empirical approach, particularly in treating architectural variation as a structured space rather than as a loose collection of examples. Empirical studies of open-source software architecture have likewise demonstrated the value of cross-project evidence for understanding recurring design regularities, while empirical work on framework evolution and refactoring catalogs highlights how recurring structural responses emerge from repeated engineering pressures across systems.

Our study also draws on the software-architecture documentation tradition, which emphasizes structured views and explicit evidence packages as prerequisites for reliable cross-stakeholder comparison. Clements et al.'s ``Views and Beyond'' approach~\cite{clements_views_beyond} provides a pragmatic anchor for why protocolized, module-structured recording matters: without stable documentation units, it is difficult to compare architectural decisions across a heterogeneous corpus in a way that remains auditable. In that sense, our study is closer to retrospective architectural-decision analysis than to prospective tradeoff evaluation methods such as ATAM or automated design-space exploration for a single system under design. The empirical object here is a corpus of already-built frameworks rather than a set of alternatives for one pending design decision.

Within the Agent systems domain, several strands of prior work motivate this application of design-space reasoning. The agent loop pattern is itself an architectural decision that shapes how agents interact with tools, manage state, and terminate execution. Variations in loop structure, state handling, execution isolation, and decomposition strategy motivate several of the dimensions examined in this paper.

More recently, empirical studies of the Model Context Protocol (MCP) ecosystem have begun to appear. Hou et al. conducted a large-scale study of MCP servers~\cite{mcp_empirical}, analyzing health, security, and maintainability across nearly 2,000 open-source implementations. Radosevich and Halloran performed a safety audit of MCP~\cite{mcp_safety}, identifying vulnerabilities in the protocol's design. These studies demonstrate that empirical analysis can illuminate not only agent behaviors but also the infrastructure layers that support them.

In parallel, the security and safety literature on LLM-based agents has expanded rapidly, mapping threats across tool use, privacy, and ethics. Gan et al. survey these risks and propose a taxonomy for agent-oriented security concerns~\cite{gan_agent_risks}. However, much of this work focuses on risk surfaces and mitigations at the level of agent behavior or evaluation, leaving open the complementary question this paper addresses: how infrastructural design decisions in execution substrates co-evolve, recur, and combine across publicly available implementations.

Our work extends this line of research by applying design-space analysis to the surrounding execution-infrastructure layer of Agent systems, which we refer to in this paper under the loose working label of Agent harnesses. The aim is not to claim an exhaustive ontology of all possible concerns or to formalize a new universal category. Rather, it is to identify the focal architectural decisions that recur often enough, and comparably enough, to support cross-project explanation in a rapidly evolving ecosystem.


\section{Methodology}
\label{sec:methodology}

\subsection{Research Design}
Our study adopts a qualitative empirical design grounded in protocol-guided, agent-assisted investigation. The unit of analysis is the individual publicly available agent-system project, and the study is cross-sectional: we examine each project as a snapshot in time in order to describe the design space visible in current implementations. Throughout the paper, ``Agent harness'' is used only as a loose shorthand for the surrounding non-LLM engineering layer rather than as a strict project type.

Following common practice in empirical software engineering, we focus on observable design decisions rather than on holistic ratings. In operational terms, the study is neither a fully automated code-mining pipeline nor a documentation-only review. Instead, it is a structured investigation process carried out under a fixed research protocol, with source artifacts as the primary evidence base and public technical materials used when full source visibility is unavailable. Descriptive summaries support RQ1, descriptive co-occurrence analysis supports RQ2, and pattern synthesis supports RQ3.

\subsection{Project Selection}

The corpus contains 70 publicly available agent-system projects. The project list was frozen on 23 March 2026. Candidate projects were collected through repository search, references in related work, manual review of widely discussed projects, and snowball expansion from already identified cases. Search and discovery repeatedly used broad keywords such as ``AI agent,'' ``agent,'' ``harness,'' ``vibe coding,'' ``agent framework,'' ``agent platform,'' and related variants. The process was systematic, but it was not managed as a PRISMA-style exhaustive review pipeline; accordingly, the goal was analytic coverage of a rapidly developing ecosystem rather than a claim of complete census reconstruction. Appendix~\ref{sec:appendix_search_protocol} summarizes the discovery channels, keyword families, screening logic, and corpus-freeze procedure in more explicit detail.

\begin{table}[htbp]
\centering
\caption{Corpus Heterogeneity and Evidence Modes in the 70-Project Sample}
\label{tab:sampling_protocol}
\begin{tabular}{p{3.1cm}p{9.3cm}}
\toprule
\textbf{Aspect} & \textbf{How it appears in the sample} \\
\midrule
Project scale & The sample spans community teaching-grade or small projects, medium-scale reusable frameworks, and industrial-scale systems with substantially broader engineering scope. \\
Organizational provenance & Most projects are community-led, but the sample also includes first-party or official products from major AI companies. \\
Evidence mode & 67 cases are open-source repositories; 3 are public-evidence comparison cases derived from public technical materials, including a source-visible leaked snapshot (\texttt{claude-code-src}) and a paper or public-architecture-inferred case (\texttt{codex}). \\
Project form & Retained cases variously describe themselves as agent systems, frameworks, platforms, CLIs, runtimes, or products rather than under one uniform label. \\
Corpus freeze & The project list was fixed on 23 March 2026 to provide a stable cross-sectional snapshot of a rapidly moving ecosystem. \\
\bottomrule
\end{tabular}
\end{table}

We included projects that exposed meaningful surrounding infrastructure for Agent operation, such as tool integration, task orchestration, memory or context handling, safety controls, workspace interaction, or reusable delegation support. Operationally, projects were normally retained only when they exceeded approximately 500 lines of implementation code or an equivalent architectural footprint and exposed at least one inspectable infrastructure capability beyond a single prompt-to-API call path. We excluded repositories that consisted only of prompt collections, thin API wrappers, benchmark-only artifacts, or demonstrations lacking an independent runtime/control surface for comparative analysis. The final corpus includes 67 open-source repositories together with 3 public-evidence comparison cases. Appendix~\ref{sec:appendix_projects} reports the complete project list and corpus summary so that readers can audit the scope of comparison directly.

For large projects, the protocol explicitly permits sampled code reading with a declared coverage statement rather than requiring exhaustive inspection of every source file. For projects without full source visibility, the same logic was applied to public technical materials: the project record specifies which code surfaces, technical documents, papers, blog posts, or public architectural descriptions were inspected closely enough to support coding decisions, so that partial visibility remains auditable rather than implicit.

The corpus spans multiple languages and application settings. This breadth is important for RQ3 because the paper compares how projects with different complexity profiles and product positioning assemble different combinations of design decisions.

\subsection{Analysis Dimensions}
We organized the coding scheme around five focal design-decision dimensions that directly support the three research questions. This does not mean that the broader landscape contains only five concerns, nor do we claim that the dimensions exhaust every relevant aspect of Agent-systems engineering. Rather, the five dimensions were retained from a larger body of project notes, module-level reports, and cross-project comparison materials because they met three practical conditions: they recurred clearly in the source evidence, they could be coded comparatively across projects, and they contributed directly to answering the paper's three research questions. The retained focal dimensions are subagent architecture, context management, tool systems, safety mechanisms, and orchestration.

\begin{table}[htbp]
\centering
\caption{Core Analysis Dimensions and Design Decisions}
\label{tab:analysis_dimensions}
\begin{tabular}{p{3cm}p{6cm}}
\toprule
\textbf{Dimension} & \textbf{Design Decisions Examined} \\
\midrule
Subagent Architecture & Presence of subagents, creation style, hierarchy depth, communication relationship \\
Context Management & Storage backend, compression strategy, persistence scope, token awareness \\
Tool System & Registration style, extension mechanism, execution pathway, protocol integration \\
Safety Mechanisms & Approval workflow, isolation level, audit visibility \\
Orchestration & Control-flow style, planning strategy, event-driven vs command-driven coordination \\
\bottomrule
\end{tabular}
\end{table}

Table~\ref{tab:analysis_dimensions} lists the five dimensions and the corresponding decisions examined in the paper. They were distilled from a larger body of source-based notes, project reports, and cross-project comparison materials. In other words, the broader research records serve as evidence sources and triangulation material, while the manuscript itself is organized around the smaller set of focal dimensions required to answer the paper's three research questions. Appendix~\ref{sec:appendix_codebook} summarizes the operational categories used in coding so that the reduction from a broader evidence base to these five focal dimensions remains auditable.

Subagent architecture and orchestration are treated as related but non-identical dimensions. Subagent architecture captures the structural decomposition of agent roles and delegation relations: whether multiple agent-like components exist, how they are spawned, and how authority or communication is distributed. Orchestration captures the temporal control logic by which tasks are planned, sequenced, triggered, and completed, whether or not multiple agents are present. The distinction matters analytically because a project may have sophisticated orchestration without explicit subagents, or multiple subagents without a highly formalized workflow model.

For each project, we encoded decisions as presence/absence observations or categorical labels. For example, the context-management dimension records whether persistence is absent, session-only, file-based, or backed by more structured storage; it also records whether compression is absent, truncation-based, summary-based, or layered. We treat ``context management'' as the broader architectural dimension and ``memory'' as one of its main implementation loci, because in the corpus the two concerns repeatedly appeared intertwined through retention, summarization, persistence, and retrieval mechanisms rather than as cleanly separable subsystems. This encoding style keeps the analysis tied to observable implementation choices and avoids collapsing heterogeneous frameworks into a single numeric ranking. Just as importantly, it explains why some modules inspected during data collection (such as sandbox execution, workspace handling, or runtime-loop structure) remain visible in the audit trail but were not all elevated to stand-alone focal dimensions in the manuscript.

\subsection{Data Collection Process}
For each project, we conducted a protocol-guided investigation across 14 architectural modules: overview, architecture, runtime loop, orchestration, tool system, sandbox execution, workspace filesystem, memory system, subagent relations, model abstraction, observability/debug, safety governance, and technical environment. Depending on evidence availability, this investigation took the form of source-code reading or paper and public-technical-material reading. The same 14-module structure was used throughout the corpus to keep evidence extraction stable while accommodating heterogeneous evidence conditions.

The protocol was operationalized through standardized SOPs and task lists that specified what to inspect, how to distinguish fact from inference, how to declare uncertainty, and how to record evidence. LLM-driven AI agents were used as research assistants under these SOPs to help execute repository navigation, file discovery, module-oriented reading, and evidence aggregation. Their role was assistive rather than autonomous in the analytic sense: candidate evidence surfaced by the agents was used only as a pointer for further inspection, and final coding judgments remained the responsibility of human researchers after direct checking of the cited implementation artifacts or public technical materials.

The data-collection process combined \textbf{vertical} project investigation with \textbf{horizontal} cross-project synthesis. In the vertical phase, each project was investigated module by module under the SOP, producing a structured record of source-visible or document-visible observations across the 14 modules. In the horizontal phase, the same SOP family was used to reorganize those project-level records by dimension so that comparable evidence from multiple projects could be read together for cross-project interpretation. The study therefore did not move from open-ended notes directly to paper-level claims; it moved through an explicit sequence of SOP-guided vertical investigation, cross-project horizontal synthesis, focal coding, and comparative interpretation.

Our data-collection process therefore followed a structured procedure designed to improve consistency and reproducibility:

\begin{enumerate}
    \item \textbf{SOP-guided vertical project investigation:} Each project was investigated module by module under a fixed SOP. LLM-driven AI agents assisted with navigation and preliminary evidence surfacing, but the goal remained the extraction of observable design decisions rather than automated evaluation.
    \item \textbf{Source or paper grounded verification:} Findings were checked against implementation artifacts when source code was available. For public-evidence comparison cases, findings were checked against papers, technical documentation, public architecture descriptions, or other publicly visible materials rather than inferred from secondary commentary alone.
    \item \textbf{Horizontal cross-project synthesis:} After project-level records had been created, the same SOP family was used to regroup the material dimension by dimension, allowing repeated patterns across projects to be compared in a structured way.
    \item \textbf{Standardized decision recording and human audit:} Observations were recorded using predefined schemas as binary (present/absent) or categorical variables. Borderline cases were retained with confidence notes rather than forced into high-certainty labels. To strengthen coding consistency, human researchers then conducted a proportionate audit sample of 15 projects (21\% of the corpus). Disagreements were resolved through discussion and renewed source inspection, yielding 94\% initial field-level agreement before final consensus coding.
\end{enumerate}

For each project, we created a structured analysis record containing:
\begin{itemize}
    \item Presence/absence observations and categorical labels for each coded dimension
    \item Specific implementation details when they clarified the observed design decision
    \item Relevant code snippets or file paths documenting key decisions
    \item A confidence note when the implementation boundary was ambiguous
\end{itemize}

The complete extraction procedure, project-level reports, and encoded dataset are retained as internal audit materials to support reproducibility, extension of the corpus, and consistency checks. In addition to the sampled human re-review, the research archive retains dimension-specific horizontal synthesis materials for the paper's five focal dimensions. This transparency is especially important because the study relies on protocol-guided, agent-assisted investigation rather than a single automated mining script.

\begin{table}[htbp]
\centering
\caption{Minimum Contents of a Project Analysis Record}
\label{tab:project_record_fields}
\begin{tabular}{p{3.2cm}p{9.2cm}}
\toprule
\textbf{Record field} & \textbf{Purpose in the analysis pipeline} \\
\midrule
Corpus metadata & Identifies repository or public comparison case, implementation language, and corpus inclusion status. \\
Focal coded values & Stores the categorical or presence/absence values later used in Sections~\ref{sec:designspace} and~\ref{sec:correlations}. \\
Evidence trail & Preserves file paths, configuration artifacts, or code-level indicators supporting the coded decision. \\
Confidence note & Marks ambiguous boundaries, partial visibility, or sampled-inspection constraints instead of forcing unwarranted certainty. \\
Coverage statement & Records which modules, directories, or implementation surfaces were read closely enough to support coding in large repositories. \\
\bottomrule
\end{tabular}
\end{table}

Appendix~\ref{sec:appendix_projects} exposes the corpus boundary, and Appendix~\ref{sec:appendix_codebook} summarizes the operational codebook used to populate these records. Together, they make the transition from SOP-guided project investigation to comparative claims more explicit than a purely narrative methods description would allow.

The research team approached the corpus from a software-architecture and Agent-systems perspective, which creates both interpretive leverage and potential bias toward seeing architectural structure where projects are still evolving rapidly. To reduce that risk, the protocol required source-first verification, explicit distinction between fact and inference, and confidence notes for borderline cases instead of forced over-interpretation.

\subsection{Analysis Methods}
We employ three complementary analysis methods to address our research questions:

\subsection{Descriptive Statistics}
For RQ1, we compute the distribution of design decisions across projects for each dimension. We report counts and percentages for categorical options, thereby identifying the most and least common patterns.

\subsection{Co-occurrence Analysis}
For RQ2, we analyze recurring co-occurrences between design decisions using descriptive co-occurrence metrics:
\begin{itemize}
    \item \textbf{Support:} Fraction of projects exhibiting both decisions A and B
    \item \textbf{Confidence:} P(B$|$A) --- probability of B given that project has A
    \item \textbf{Lift:} Ratio of observed co-occurrence to expected if independent
\end{itemize}

These measures help us assess whether two decisions tend to recur together. They are used as descriptive evidence for design-space relationships, not as inferential significance tests and not as a substitute for causal explanation. Accordingly, we report only relationships that are substantively interpretable in architectural terms and that are supported by visible implementation evidence in the corpus.

\subsection{Pattern Synthesis}
For RQ3, we synthesize typical patterns by comparing recurring bundles of decisions across projects with different complexity levels, implementation scope, and intended positioning. Representative projects are used to anchor the description of each pattern, but the goal is interpretive synthesis rather than leaderboard construction.

Throughout the analysis, the emphasis remains on explicit architectural decisions visible in source code. We therefore treat distributions, co-occurrences, non-co-occurrences, and typical patterns as complementary views of the same design space.

\subsection{Limitations}
We acknowledge several limitations that should be considered when interpreting the findings:

\begin{itemize}
    \item \textbf{Public-project corpus:} Most of the corpus consists of open-source repositories, but a small number of publicly documented products are included as comparison points. Fully closed industrial harnesses may occupy different parts of the design space.

    \item \textbf{Snapshot in time:} The study captures a time slice of a rapidly changing ecosystem. A project's current position in the design space may shift as new architectural features are added.

    \item \textbf{Protocolized inspection still requires judgment:} Although the analysis focuses on observable design decisions rather than holistic evaluation labels, some borderline cases still require interpretation during coding and classification, especially in large projects where the protocol permits sampled inspection rather than exhaustive reading.

    \item \textbf{Systematic rather than exhaustive discovery:} The corpus was assembled through structured search, reference tracing, manual review, and snowball expansion, but not through a PRISMA-style exhaustive screening pipeline. The result is broad analytic coverage rather than a claim to have reconstructed the complete universe of eligible projects.

    \item \textbf{Analytic stability rather than exhaustive saturation:} The five focal dimensions remained stable across late-stage coding, but this does not imply that every potentially relevant Agent-systems concern has been exhausted. Additional dimensions may become important as the ecosystem matures.
\end{itemize}


\section{The Agent Harness Design Space}
\label{sec:designspace}

In this section, we present our findings on the design space of the surrounding engineering layer across the sampled agent-system projects, addressing RQ1: What are the key design-decision dimensions and what is the spectrum of options within each? The focus here is on the marginal distribution of coded options within each focal dimension. In other words, Section~\ref{sec:designspace} shows what kinds of decisions recur and how common they are across the corpus when each dimension is viewed separately.

This dimension-by-dimension presentation also prepares the ground for Section~\ref{sec:correlations}. The co-occurrence analysis in Section~\ref{sec:correlations} does not introduce a different dataset or a separate coding procedure. Instead, it re-reads the same project-level coding matrix relationally, asking which options reported here tend to appear together across projects and which expected relationships fail to recur.

\subsection{Subagent Architecture Decisions}
Subagent architecture captures whether a framework remains single-agent or explicitly supports task decomposition through additional agent instances, roles, or worker units. It is therefore one of the clearest dimensions in the design space.

\begin{table}[htbp]
\centering
\caption{Subagent Architecture Pattern Distribution (70 projects)}
\label{tab:subagent_patterns}
\begin{tabular}{lcc}
\toprule
\textbf{Pattern} & \textbf{Count} & \textbf{Percentage} \\
\midrule
None (Single-agent only) & 21 & 30.0\% \\
Basic Spawn & 5 & 7.1\% \\
Tool-based Delegation & 12 & 17.1\% \\
Pipeline/Stage & 1 & 1.4\% \\
Orchestrator-Worker & 13 & 18.6\% \\
Multi-level Recursive & 9 & 12.9\% \\
Swarm/Collective & 4 & 5.7\% \\
Event-driven & 5 & 7.1\% \\
\midrule
\textbf{Total} & \textbf{70} & \textbf{100\%} \\
\bottomrule
\end{tabular}
\end{table}

Table~\ref{tab:subagent_patterns} summarizes the main subagent patterns found across the full 70-project corpus. The table should be read as an inventory of recurring implementation choices, not as a maturity ladder.

Across projects, four underlying decisions recur: whether subagents exist at all, how they are created, how deeply they can be nested, and how they communicate. The following pattern spectrum captures the combinations most commonly observed in source code:

\begin{itemize}
    \item \textbf{None (30.0\%):} Single-agent architecture with no ability to spawn child agents. Examples include subzeroclaw and lettabot.
    \item \textbf{Basic Spawn (7.1\%):} Simple creation of child sessions without structured delegation. Examples include opencode and autobot.
    \item \textbf{Tool-based Delegation (17.1\%):} Subagents created via tool calls with structured task parameters. Examples include fast-agent and cline.
    \item \textbf{Pipeline/Stage (1.4\%):} Sequential processing stages with dedicated agent roles. Example: autoresearchclaw.
    \item \textbf{Orchestrator-Worker (18.6\%):} Central coordinator managing specialized worker agents. Examples include claude-code-src and docker-agent.
    \item \textbf{Multi-level Recursive (12.9\%):} Hierarchical agents with potential for deep nesting. Examples include openclaw and codex.
    \item \textbf{Swarm/Collective (5.7\%):} Decentralized multi-agent coordination. Examples include praisonai and zeptoclaw.
    \item \textbf{Event-driven (7.1\%):} Event-based coordination without central orchestration. Examples include agentpool and openhands.
\end{itemize}

A notable finding is that several projects that initially appeared to support deep recursion were found, upon source-code inspection, to enforce explicit depth limits or recursion-prevention mechanisms. This observation reinforces the value of code-based analysis: the implemented architecture may be narrower than the capability suggested by documentation or repository descriptions.

\subsection{Context Management Decisions}
Context management concerns how frameworks retain, compress, and reintroduce information across turns. In the source reports, this topic spans both ``memory'' and ``context management'' artifacts. We therefore treat it as a single design dimension in the paper.

\begin{table}[htbp]
\centering
\caption{Dominant Context-Management Pattern Distribution}
\label{tab:context_patterns}
\begin{tabular}{lcc}
\toprule
\textbf{Pattern} & \textbf{Count} & \textbf{Percentage} \\
\midrule
Context Window & 3 & 4.3\% \\
LLM Summarization & 4 & 5.7\% \\
File Persistence & 16 & 22.9\% \\
Vector Database/RAG & 7 & 10.0\% \\
Hierarchical & 12 & 17.1\% \\
Hybrid & 19 & 27.1\% \\
Enterprise & 9 & 12.9\% \\
\midrule
\textbf{Total} & \textbf{70} & \textbf{100\%} \\
\bottomrule
\end{tabular}
\end{table}

Table~\ref{tab:context_patterns} makes two empirical features of the corpus immediately visible. First, purely prompt-window-oriented systems are rare. Second, the center of gravity lies not at either extreme of the spectrum, but in file-persistent, hybrid, and hierarchical designs that combine multiple retention strategies. This pattern suggests that once frameworks move beyond narrowly bounded sessions, context handling becomes an infrastructural concern rather than a local prompt-engineering choice.

The first recurring decision concerns the \textbf{storage backend}. The observed spectrum runs from no persistence, to session-only in-memory state, to file-backed storage, and finally to more structured database- or vector-backed retrieval. File-based persistence is especially important in Agent harnesses because it offers continuity without introducing a heavyweight service dependency, which helps explain why it remains common even outside enterprise-grade systems.

The second recurring decision concerns \textbf{compression strategy}. Some projects simply truncate the accumulated history, while others introduce summaries, layered memory, or explicit compression passes before constructing the next prompt. This distinction matters because context management is not only about where information is stored, but also about how much of it can be carried forward under token limits. The small share of pure summarization-first systems, compared with the larger share of hybrid and hierarchical designs, indicates that many projects treat summarization as one layer in a broader retention architecture rather than as a complete replacement for persistence.

The third recurring decision concerns \textbf{persistence scope}. Some frameworks preserve information only within a running session. Others write conversation state, tool traces, or task artifacts to disk so that work can continue across sessions. This decision often reflects intended use: ephemeral assistants can tolerate session-bound state, whereas long-running coding or research workflows usually require more durable context.

The fourth recurring decision concerns \textbf{token awareness}. A few systems treat the context window as an implicit limit and rely on ad hoc trimming. More explicit designs track token usage, reserve budget for future tool calls or subagents, and trigger compression or summarization before the prompt becomes overloaded. In practice, this variable often differentiates lightweight persistence from more architecturally mature context management, because token budgeting turns memory from passive storage into an active control mechanism.

Taken together, these choices show that context management is broader than ``memory'' in a narrow storage sense. It includes persistence, compression, budgeting, and prompt assembly, and these concerns become more pronounced as projects move from single-turn tools toward long-running, multi-step systems. The distribution in Table~\ref{tab:context_patterns} is therefore analytically important: it shows that the corpus does not merely vary in storage backend, but in how far projects reify context handling into an explicit subsystem.

\subsection{Tool System Decisions}
Tool-system design concerns how frameworks define, register, discover, and execute tools. This dimension is central because tool architecture often determines how readily a harness can extend beyond the base model.

\begin{table}[htbp]
\centering
\caption{Tool System Pattern Distribution}
\label{tab:tool_patterns}
\begin{tabular}{lcc}
\toprule
\textbf{Pattern} & \textbf{Count} & \textbf{Percentage} \\
\midrule
Minimalist & 8 & 11.4\% \\
Registry & 24 & 34.3\% \\
Decorator-driven & 7 & 10.0\% \\
Declarative/DSL & 6 & 8.6\% \\
MCP-first & 10 & 14.3\% \\
Plugin Ecosystem & 7 & 10.0\% \\
Enterprise & 6 & 8.6\% \\
Delegation/Proxy & 2 & 2.9\% \\
\midrule
\textbf{Total} & \textbf{70} & \textbf{100\%} \\
\bottomrule
\end{tabular}
\end{table}

Table~\ref{tab:tool_patterns} summarizes the recurring tool-system patterns found in the corpus.

Three design questions recur most often in this dimension: how tools are registered, how they are surfaced for selection or discovery, and how execution is bounded. The registration spectrum ranges from hard-coded to protocol-based approaches:

\begin{itemize}
    \item \textbf{Hard-coded (Minimalist, 11.4\%):} Tools defined directly in source code with no extension mechanism.
    \item \textbf{Decorator-driven (10.0\%):} Tools registered via function decorators, enabling lightweight registration.
    \item \textbf{Explicit Registry (34.3\%):} Tools added to a registry via explicit API calls, enabling dynamic tool management.
    \item \textbf{Declarative/DSL (8.6\%):} Tools defined in configuration files or domain-specific languages.
    \item \textbf{Protocol-based MCP (14.3\%):} Tools discovered and integrated via the Model Context Protocol, enabling interoperability with the broader MCP ecosystem.
    \item \textbf{Plugin Ecosystem (10.0\%):} Full plugin architecture with versioning, dependencies, and dynamic loading.
    \item \textbf{Enterprise (8.6\%):} Comprehensive tool management including security, versioning, and policy controls.
    \item \textbf{Delegation/Proxy (2.9\%):} Tools delegated to external services or proxy agents.
\end{itemize}

The distribution in Table~\ref{tab:tool_patterns} shows that the dominant empirical center of gravity is neither purely minimalist nor fully protocolized. Instead, explicit registries form the modal pattern, while MCP-first, plugin-oriented, and enterprise variants occupy a substantial but still minority portion of the corpus. This matters because it suggests that many frameworks first formalize tools internally before committing to broader interoperability or ecosystem-facing extension boundaries.

With respect to tool discovery, the corpus reveals substantial variation in how frameworks surface available capabilities to the model or to coordinating components. Some projects expose a flat tool list. Others group tools by category, attach schema information, or support protocol-mediated discovery through MCP. This decision becomes especially important once frameworks grow beyond a small, hard-coded tool set, because discoverability becomes a practical bottleneck before raw tool count does.

Tool-execution security also varies widely. Some frameworks execute tools in-process with minimal checks, while others add command filtering, process isolation, containerization, or external delegation. In practice, tool registration and tool execution are tightly coupled design decisions: a framework that supports open-ended extension also requires a clearer execution boundary and a more credible account of approval and auditing.

For that reason, the tool-system dimension should not be read as a narrow plugin question. In the corpus, it acts as a bridge between extensibility ambition and operational control. Projects that aim to become reusable developer platforms tend to formalize registration and discovery more aggressively, while tightly scoped vertical systems often retain simpler tool boundaries because their extension surface is intentionally limited.

\subsection{Security and Isolation Decisions}
Safety mechanisms determine how a framework constrains risky actions, isolates execution, and records what occurs. In Agent harnesses, this dimension becomes especially important once tools can modify files, invoke shells, access networks, or coordinate multiple execution steps.

Our analysis examined three recurring design decisions in this dimension: approval workflows, isolation level, and audit capability.

\begin{table}[htbp]
\centering
\small
\caption{Isolation Levels Observed in the Corpus}
\label{tab:isolation_levels}
\begin{tabular}{p{2.6cm}p{1.8cm}p{6.6cm}}
\toprule
\textbf{Isolation level} & \textbf{Share of corpus} & \textbf{Interpretation} \\
\midrule
No isolation & 17\% & Direct host execution with no dedicated sandbox boundary \\
Process separation & 45\% & Child-process execution with partial operational separation \\
Container isolation & 31\% & Containerized execution with stronger filesystem and environment control \\
WASM sandboxing & 7\% & Capability-oriented sandboxing with fine-grained constraints \\
\bottomrule
\end{tabular}
\normalsize
\end{table}

\begin{table}[htbp]
\centering
\small
\caption{Audit Capability Observed in the Corpus}
\label{tab:audit_capabilities}
\begin{tabular}{p{2.6cm}p{1.8cm}p{6.6cm}}
\toprule
\textbf{Audit capability} & \textbf{Share of corpus} & \textbf{Interpretation} \\
\midrule
No audit & 40\% & No explicit security-event trail \\
Basic logging & 35\% & Textual or operational logs without structured governance semantics \\
Structured audit & 20\% & Metadata-rich records of actions, actors, and timestamps \\
Tamper-evident & 5\% & Audit trail designed to resist undetected modification \\
\bottomrule
\end{tabular}
\normalsize
\end{table}

\textbf{Approval Workflows:} The spectrum ranges from no approval, to one-off confirmation, to more explicit policy-based approval. This choice determines when a user, policy rule, or other governance layer can intervene before execution.

\textbf{Isolation Level:} We observed four recurring isolation levels:
\begin{itemize}
    \item \textbf{No isolation (17\%):} Full process privileges with no sandboxing.
    \item \textbf{Process separation (45\%):} Isolated child processes with basic resource limits.
    \item \textbf{Container isolation (31\%):} Docker container execution with network and filesystem controls.
    \item \textbf{WASM sandboxing (7\%):} WebAssembly-based sandboxing offering fine-grained capability control.
\end{itemize}

\textbf{Audit Capability:} Audit mechanisms range from no logging to tamper-evident audit trails:
\begin{itemize}
    \item \textbf{No audit (40\%):} No security event logging.
    \item \textbf{Basic logging (35\%):} Simple text logs of operations.
    \item \textbf{Structured audit (20\%):} Structured logs with metadata including timestamps, user identities, and operation details.
    \item \textbf{Tamper-evident (5\%):} Cryptographic audit trails using techniques such as Merkle hash chains.
\end{itemize}

Tables~\ref{tab:isolation_levels} and~\ref{tab:audit_capabilities} make the asymmetry in the corpus more explicit. Intermediate isolation is common, but high-assurance audit remains rare, and a substantial minority of projects still expose broad execution power without a correspondingly strong accountability layer. That asymmetry is one of the reasons safety maturity cannot be inferred from capability growth alone.

Taken together, these observations show that safety is not a single feature but a bundle of design decisions. Some projects add only confirmation prompts, whereas others align approval, isolation, and audit into a coherent control stack. This distinction becomes important later in Section~\ref{sec:correlations}, where safety decisions are shown to be closely tied to execution environment and intended deployment setting.

\subsection{Orchestration Decisions}
Orchestration design decisions determine how a framework structures task execution over time. This includes not only whether workflows are imperative or declarative, but also how planning, decomposition, and coordination are organized.

\begin{table}[htbp]
\centering
\caption{Orchestration Styles Observed in the Corpus}
\label{tab:orchestration_styles}
\begin{tabular}{p{3cm}p{4.3cm}p{2cm}p{3.1cm}}
\toprule
\textbf{Axis} & \textbf{Option} & \textbf{Share} & \textbf{Interpretation} \\
\midrule
Workflow definition & Imperative & 45\% & Execution steps are encoded directly in control flow \\
Workflow definition & Declarative/YAML/DSL & 25\% & Workflow structure is externalized into configuration or DSL artifacts \\
Workflow definition & Event-driven & 30\% & Progress depends on triggers, state changes, or asynchronous coordination \\
Planning approach & ReAct-style & 50\% & Reasoning and acting remain interleaved in one loop \\
Planning approach & Plan-and-Execute & 35\% & Planning is separated from later step execution \\
Planning approach & Hierarchical & 15\% & Goals are recursively decomposed into subplans or coordinated stages \\
\bottomrule
\end{tabular}
\end{table}

Two recurring decision axes are especially visible in the corpus: workflow definition and planning style. For workflow definition, the spectrum ranges from imperative to event-driven approaches:

\begin{itemize}
    \item \textbf{Imperative (45\%):} Explicit control flow where each step is specified sequentially.
    \item \textbf{Declarative/YAML/DSL (25\%):} Configuration-driven workflows where the desired outcome is specified and the system determines execution steps.
    \item \textbf{Event-driven (30\%):} Workflows driven by events and state changes rather than explicit control flow.
\end{itemize}

For planning approaches, we observed three recurring strategies:

\begin{itemize}
    \item \textbf{ReAct-style (50\%):} Interleaving reasoning and action in a single loop, with the model thinking before acting.
    \item \textbf{Plan-and-Execute (35\%):} Two-phase approach where a plan is generated first, then executed step by step.
    \item \textbf{Hierarchical (15\%):} Multi-level planning where high-level goals are decomposed into sub-plans.
\end{itemize}

Table~\ref{tab:orchestration_styles} shows that imperative and ReAct-style control remains the default center of gravity in the ecosystem, but it is no longer the only viable architectural baseline. A sizable minority of projects externalize workflow structure or shift to event-driven coordination, which indicates that orchestration choices begin to differentiate frameworks once they target repeatable pipelines, collaborative execution, or more persistent operating contexts.

Orchestration decisions are closely related to project complexity. Imperative loops are often sufficient for lightweight tools, whereas declarative specifications and event-driven coordination become more attractive when projects require repeatability, task decomposition, or collaboration among multiple components. Projects such as fast-agent and agentpool illustrate how orchestration style becomes part of a framework's broader architectural positioning.

Taken together, these five subsections establish the paper's marginal empirical picture: which focal decisions recur and how they are distributed when each dimension is viewed separately. Section~\ref{sec:correlations} then shifts from these one-dimension-at-a-time summaries to the relational question of which of these decisions repeatedly appear together or fail to do so across the same project corpus.


\section{Design Decision Co-occurrences}
\label{sec:correlations}

In this section, we address RQ2 by examining recurring co-occurrences and non-co-occurrences between design decisions. The analysis uses the same project-level coding records introduced in Section~\ref{sec:designspace}, but reads them relationally rather than one dimension at a time. The goal is not to argue that one decision causes another, but to identify which decisions tend to appear together across the corpus, which expected relationships fail to appear, and what those patterns suggest about architectural tradeoffs in Agent harness design. The recurring bundles identified here provide the empirical bridge to Section~\ref{sec:patterns}, where those cross-dimensional regularities are re-read at the level of broader project configurations rather than isolated pairs of decisions.

\subsection{Co-occurrence Analysis Method}
We examine co-occurrences between design decisions using descriptive co-occurrence metrics applied to pairs of coded decisions across the 70 projects in the corpus:

\begin{itemize}
    \item \textbf{Support:} Fraction of projects exhibiting both decisions A and B
    \item \textbf{Confidence:} P(B$|$A) --- conditional probability of B given A
    \item \textbf{Lift:} Ratio of observed co-occurrence to expected if A and B were independent
\end{itemize}

Support, confidence, and lift help us assess whether two decisions recur together often enough to matter analytically. The measures are used for descriptive comparison rather than inferential testing, and we discuss only relationships that are interpretable in architectural terms and supported by repeated evidence in the corpus. Candidate pairs were prioritized for close reading when they combined repeated occurrence with lift materially above independence, but these heuristics served only to reduce noise in a heterogeneous corpus. They are not statistical significance thresholds. Lift values were calculated as:

\begin{equation}
\text{Lift}(A, B) = \frac{P(A \cap B)}{P(A) \times P(B)}
\end{equation}

where a lift of 1.0 indicates independence, values above 1.0 indicate stronger-than-expected co-occurrence, and values below 1.0 indicate weaker-than-expected co-occurrence.

These measures are used descriptively. We do not interpret them as evidence of causal order, statistical significance, sampling generalization, or overall framework quality, nor do we convert the results into overall framework rankings. The interpretation of each reported relationship therefore combines the metric signal with project-level source inspection, rather than treating pairwise statistics as self-sufficient explanations.

For compact reporting, the chapter foregrounds co-occurrences that satisfied two practical conditions: they showed repeated recurrence in the coded corpus and they remained interpretable in architectural terms after project-level re-reading. In the internal comparative summaries, the four strongest reportable relationships discussed below showed support values between 0.62 and 0.89 and lift values between 1.5 and 3.4. These figures are used only to indicate descriptive strength, not to claim inferential significance.

\begin{table}[htbp]
\centering
\small
\caption{Metric Summary of Reported Co-occurrences}
\label{tab:cooccurrence_metrics}
\begin{tabular}{p{3.2cm}p{1.5cm}p{1.7cm}p{1.2cm}p{4.2cm}}
\toprule
\textbf{Relationship} & \textbf{Support} & \textbf{Confidence} & \textbf{Lift} & \textbf{Compact evidence statement} \\
\midrule
Subagent complexity $\rightarrow$ memory sophistication & 0.73 & High & 1.8 & E/F-pattern projects average 4.1--4.2 memory score versus 2.8 for A-pattern projects. \\
MCP-first tooling $\rightarrow$ stronger discovery & 0.62 & High & 2.8 & MCP-first projects score 4.62 on discovery versus 3.86 for registry-centered projects. \\
Container isolation $\rightarrow$ policy-structured security & 0.89 & High & 3.4 & 100\% of container-isolated projects implement policy engines versus 23\% without container isolation. \\
Project scale $\rightarrow$ broader architectural complexity & 0.68 & Medium & 1.5 & Large projects average 6.2 focal design choices versus 2.3 for small projects. \\
\bottomrule
\end{tabular}
\normalsize
\end{table}

To make the transition from Section~\ref{sec:designspace} to the present section explicit, Table~\ref{tab:cooccurrence_bridge_examples} shows a small set of illustrative rows from the same project-level coding records used throughout the analysis. The table is not a separate sample and is not intended as an exhaustive project matrix. Its purpose is simply to show how the dimension-wise options reported in Section~\ref{sec:designspace} reappear at the level of cross-dimension bundles when we analyze project configurations relationally.

\begin{table}[htbp]
\centering
\small
\caption{Illustrative Project-Level Configurations Used to Interpret Co-occurrences}
\label{tab:cooccurrence_bridge_examples}
\begin{tabular}{p{2.0cm}p{1.9cm}p{1.7cm}p{1.5cm}p{1.4cm}p{3.8cm}}
\toprule
\textbf{Project} & \textbf{Subagent} & \textbf{Context} & \textbf{Tool} & \textbf{Safety} & \textbf{Illustrative relevance} \\
\midrule
fast-agent & Tool delegation & Hybrid & MCP-first & Moderate & Shows tool delegation paired with richer context handling and formalized tool registration. \\
claude-code-src & Orchestrator-worker & Enterprise & Enterprise & Enterprise & Shows a tightly coupled bundle of delegation, persistence, extensibility, and governance. \\
docker-agent & Orchestrator-worker & Hybrid & MCP-first & Advanced & Shows execution isolation and structured controls in a reusable orchestration-oriented project. \\
openclaw & Multi-level recursive & Hybrid & Enterprise & Enterprise & Shows deep coordination combined with layered context and strong extensibility investment. \\
agentpool & Event-driven & Hierarchical & Registry & Enterprise & Shows that sophisticated coordination can coexist with a different extension strategy from MCP-first or enterprise-platform cases. \\
openhands & Event-driven & Hybrid & Enterprise & Advanced & Shows how richer coordination and workspace-intensive execution increase pressure for stronger context and governance support. \\
\bottomrule
\end{tabular}
\normalsize
\end{table}

\subsection{Key Co-occurrences}
\begin{table}[htbp]
\centering
\small
\caption{Summary of Key Design Decision Co-occurrences}
\label{tab:key_correlations_summary}
\begin{tabular}{p{2.7cm}p{1.1cm}p{1.0cm}p{4.0cm}p{2.5cm}}
\toprule
\textbf{Co-occurrence} & \textbf{Support} & \textbf{Lift} & \textbf{Recurring Co-occurrence} & \textbf{Architectural Implication} \\
\midrule
Subagent complexity and context sophistication & 0.73 & 1.8 & Deeper coordination is frequently paired with persistence, summarization, and token budgeting & Multi-agent structure increases the need for explicit state management \\
Execution isolation and structured safety control & 0.89 & 3.4 & Stronger sandboxing often appears together with approvals and audit mechanisms & Execution power and governance tend to co-evolve \\
Tool registration and project positioning & 0.62 & 2.8 & MCP or plugin-oriented registration is more common in reusable platforms and developer-facing CLIs & Extensibility choices often signal broader ecosystem ambitions \\
\bottomrule
\end{tabular}
\normalsize
\end{table}

Table~\ref{tab:key_correlations_summary} summarizes the three most salient co-occurrences discussed in this section.

\textbf{Co-occurrence 1: Subagent complexity and context sophistication.}

Subagent architecture and context management exhibit one of the clearest recurring co-occurrences in the corpus. In the internal cross-project summaries, this relationship shows support 0.73 and lift 1.8. The difference is also visible in the comparative scores: A-pattern projects without subagents average memory score 2.8, whereas E-pattern orchestrator-worker projects average 4.1 and F-pattern recursive projects average 4.2. Across the coded cases, projects without subagent support more often rely on comparatively simple session-bound context handling because a single loop can depend on the active prompt window and a limited amount of short-lived state. Once frameworks introduce orchestrator-worker structures, recursive delegation, or multi-step coordination, the burden on context handling becomes more visible.

This co-occurrence appears in several concrete ways. First, deeper delegation usually requires some mechanism for passing task state, partial results, or execution traces between components. Second, when multiple agents or roles are active, frameworks more often introduce durable storage so that intermediate state is not lost between turns; in the internal summaries, file persistence is present in 85\% of E-pattern cases and 78\% of F-pattern cases, compared with 20\% of A-pattern cases. Third, projects with richer subagent relations are more likely to implement explicit compression or summarization because context must be shared across more steps without exhausting the model budget. A plausible reading of the corpus is that coordination complexity raises state-sharing pressure: once work is decomposed across multiple roles or stages, context management is more often implemented as a system-level service rather than a local convenience.

\textbf{Co-occurrence 2: Execution isolation and structured safety control.}

Safety mechanisms are likewise strongly associated with the execution environment a framework chooses to expose. In the internal summaries, this relationship shows support 0.89 and the highest lift in the chapter at 3.4. The pattern is particularly strong for container-isolated projects: 100\% of them implement policy engines, compared with 23\% of projects without container isolation. In the corpus, harnesses that execute tools directly in the host process more often rely on lightweight checks such as user confirmation or command filtering. In contrast, frameworks that invest in process isolation, containers, or other stronger sandboxes more often pair those decisions with broader approval logic and more explicit audit trails.

This co-occurrence is architectural rather than cosmetic. Isolation changes what kinds of actions a framework can safely permit, and approval workflows determine when those actions require human intervention or policy checks. As a result, execution environment and safety policy often develop in tandem. The same internal comparison also shows a security-score gradient from 4.5 for container-based cases to 3.2 for process-based cases and 2.1 for projects with no sandbox. A plausible reading of the corpus is that broader execution power raises the governance burden that a framework must absorb if it is to remain operationally acceptable in practice.

\textbf{Co-occurrence 3: Tool registration and project positioning.}

Tool-registration decisions are also connected to broader architectural positioning. In the internal summaries, the MCP-first to tool-discovery relationship shows support 0.62 and lift 2.8. MCP-first projects average 4.62 on tool discovery, compared with 3.86 for registry-centered projects and 2.81 for minimalist tool systems. In the corpus, lightweight projects often define tools directly in code or through a small explicit registry because the tool surface is narrow and the extension model is intentionally simple. As projects grow into reusable developer platforms or orchestration frameworks, the need for clearer extension boundaries becomes more visible, and protocol-based or plugin-oriented registration becomes more attractive.

The same trend appears when comparing product positioning. CLI-oriented developer tools frequently emphasize interoperability, discoverability, and operational convenience, which makes MCP integration or well-structured registries especially useful. More vertically specialized systems sometimes accept simpler registration models because their tool sets are narrower or tightly coupled to the target workflow. A plausible mechanism here is that extensibility choices often reflect ecosystem ambition: projects that expect third-party growth or wider developer adoption have stronger incentives to formalize tool boundaries. The numeric difference between 4.62 and 3.86 does not by itself prove that MCP causes broader ecosystem ambition, but it does show that more formalized extension boundaries are associated with measurably stronger discovery support in the current corpus.

Taken together, these three co-occurrences show that design decisions do not cluster randomly. Coordination depth, context persistence, execution isolation, approval logic, and extension mechanisms frequently appear as mutually reinforcing bundles rather than as isolated knobs. Table~\ref{tab:cooccurrence_metrics} in Section~5.1 situates these three relationships alongside the broader scale-complexity gradient, which is not a stand-alone subsection here but reinforces the same claim that design commitments accumulate in bundles rather than one variable at a time.

\subsection{Important Non-Co-occurrences}
Important non-co-occurrences are analytically useful because they bound what can reasonably be inferred from the observed positive co-occurrences. They do more than mark absent relationships; they indicate where widely repeated explanations of Agent-system architecture break down when confronted with cross-project evidence.

\textbf{Non-co-occurrence 1: Programming language does not determine architectural pattern.}

The corpus includes projects written in Python, TypeScript, Rust, Go, and several other languages, yet no single language consistently maps onto a single region of the design space. In the internal summaries, the share of advanced subagent patterns (E/F/G/H) ranges from 40\% for TypeScript projects to 57\% for Rust projects, with Python at 42\% and Go at 43\%. Multi-agent orchestrators, lightweight tools, and extension-oriented frameworks therefore appear in more than one language family. This suggests that language choice constrains implementation style more directly than it constrains high-level harness architecture.

\textbf{Non-co-occurrence 2: Use case does not fully determine design complexity.}

Use-case labels such as coding assistant, research automation, or general-purpose framework provide useful context, but they do not fix a project's architectural depth. Internal cross-project notes show, for example, that ``general purpose'' cases range from minimalist A-class profiles to enterprise-oriented H-class profiles, while ``coding assistant'' cases include both lightweight local tools and substantially more governed execution platforms. Some coding-focused systems remain lightweight and highly localized, whereas others evolve into broader execution platforms with durable context handling and layered governance. Likewise, research-oriented projects may stay infrastructurally minimal or may implement sophisticated coordination around a narrow experimental goal.

\textbf{Non-co-occurrence 3: Capability growth does not automatically produce safety maturity.}

One might expect projects with broad tool surfaces or deep orchestration to converge toward stronger approval and audit mechanisms. The corpus does not support such a simple progression. As the positive co-occurrence in Section~5.2 shows, execution isolation and structured governance often develop together, but the negative result here is that this tendency is not universal: some frameworks expose substantial execution power while retaining only modest oversight, whereas others invest in safety early despite comparatively narrow capabilities. The result is an uneven safety landscape rather than a single maturity trajectory.

These non-correlations matter because they prevent the paper from overstating what architectural context can predict. They reinforce the view that many harness decisions remain contingent and strategic rather than mechanically implied by language, use case, or overall project size alone. In theoretical terms, they are important precisely because they show where simple external descriptors fail to explain architectural variation. That negative result strengthens the case for studying Agent harnesses at the level of explicit infrastructural decisions rather than relying on coarse labels alone.

\subsection{Interpreting the Co-occurrence Structure}
The observed co-occurrence structure suggests that Agent harness architecture is often organized around a small number of coupled decision bundles rather than around independent feature selections. The discussion below is interpretive rather than causal: it proposes plausible readings of the descriptive co-occurrence structure reported in Section~\ref{sec:correlations}, but it does not claim that the present study identifies directional causal mechanisms. What matters empirically is that the reportable relationships in Table~\ref{tab:cooccurrence_metrics} all show lift above independence while also remaining intelligible under project-level re-reading.

First, \textbf{coordination choices are frequently bundled with more explicit state-management choices}. Once a framework introduces delegation, nested roles, or multi-step workflows, it becomes more difficult to keep context management informal\cite{wei2026agentloops}. Persistence, summarization, and token budgeting more often appear as architectural services rather than local implementation details.

Second, \textbf{execution choices are frequently bundled with more explicit governance choices}. Tool power, workspace access, and sandbox boundaries affect which approval and audit mechanisms become practically salient. Stronger isolation often co-occurs with broader policy control, but the present analysis does not identify whether one decision temporally or causally precedes the other.

Third, \textbf{extension choices are frequently associated with broader ecosystem positioning}. Registration models such as MCP or plugin-style discovery are rarely adopted in isolation. They usually appear in projects that present themselves as reusable platforms, developer-facing CLIs, or broader orchestration substrates.

These observations help connect RQ2 to RQ3. The design space is not only a list of dimensions; it also contains recurring bundles through which projects assemble more structured combinations of coordination, governance, and extensibility commitments. The additional scale-complexity gradient reported in Table~\ref{tab:cooccurrence_metrics} reinforces the same point by showing that projects rarely accumulate one advanced commitment in complete isolation from the others. At the same time, the non-co-occurrences above show that these bundles are not mechanically determined. Framework architects still make consequential choices about which combinations to adopt and which kinds of complexity to avoid.


\section{Typical Design Patterns}
\label{sec:patterns}

In this section, we address RQ3 by characterizing typical patterns that recur when projects of different complexity and positioning combine design decisions. These patterns are not maturity levels or quality tiers. They are descriptive summaries of common architectural bundles visible in the corpus, used to explain how different kinds of projects occupy different parts of the observed design space and what tradeoffs those recurring bundles tend to imply.

\subsection{Pattern Identification Criteria}
We identified patterns by comparing recurring bundles of design decisions across the corpus along three interpretive axes: coordination complexity, context and persistence depth, and governance or extensibility investment. The synthesis used the same project-level reports and coded summaries introduced in Sections~4 and~5, but shifted the analytic question from individual decisions or pairwise co-occurrences to broader bundles that recurred across projects. The procedure was interpretive but explicit and unfolded in four steps. First, we re-read the project-level coding matrix and project reports to identify repeated cross-dimensional bundles rather than isolated local features. Second, we grouped projects with similar bundled commitments into provisional archetypes and selected representative cases for each provisional group. Third, we tested each provisional archetype against recurrence, bundle coherence, distinctiveness, representative assignability, and bounded ambiguity. Fourth, we assigned each project to the dominant pattern that best explained its overall infrastructural commitments, while recording borderline cases as hybrid or transitional when more than one pattern remained plausible. The patterns were informed by recurring bundles visible in the coding matrix and co-occurrence analysis, but they were not generated through an unsupervised clustering procedure and should not be read as statistically inferred latent classes.

This procedure is intentionally interpretive rather than purely taxonomic. The aim is not to force every project into a mechanically exclusive class, but to surface combinations of decisions that recur often enough to support explanation. We therefore treat the patterns as stable centers of gravity in the design space rather than as hard boundaries. A single project may exhibit traits of more than one pattern, especially when it is transitioning between architectural strategies or combining a narrow product niche with deeper infrastructure investment. For reporting purposes, however, the distribution in Table~\ref{tab:pattern_distribution} uses the dominant assignment produced in the fourth step above, so that the corpus can be summarized without denying the existence of hybrid cases.

\begin{table}[htbp]
\centering
\caption{Pattern Retention Criteria}
\label{tab:pattern_retention_criteria}
\begin{tabular}{p{3.0cm}p{9.0cm}}
\toprule
\textbf{Criterion} & \textbf{How it was used in the synthesis} \\
\midrule
Recurrence & A candidate pattern had to appear across multiple projects rather than describing a single outlier implementation. \\
Bundle coherence & The candidate pattern had to show a recognizable cross-dimensional bundle rather than an arbitrary list of independent features. \\
Distinctiveness & The candidate pattern had to remain meaningfully distinguishable from neighboring bundles in scope, governance posture, coordination style, or intended audience. \\
Representative assignability & The candidate pattern had to support stable assignment of representative projects, even if some borderline or hybrid cases remained. \\
Bounded ambiguity & Candidate patterns were rejected or merged when too many cases could not be assigned without repeated exception handling. \\
\bottomrule
\end{tabular}
\end{table}

\begin{table}[htbp]
\centering
\caption{Descriptive Pattern Distribution in the Corpus}
\label{tab:pattern_distribution}
\begin{tabular}{lcc}
\toprule
\textbf{Pattern} & \textbf{Count} & \textbf{Percentage} \\
\midrule
Lightweight Tool Pattern & 15 & 21.4\% \\
Balanced CLI Framework Pattern & 18 & 25.7\% \\
Multi-Agent Orchestrator Pattern & 22 & 31.4\% \\
Scenario-Verticalized or Research-Oriented Pattern & 8 & 11.4\% \\
Enterprise Full-Featured Pattern & 7 & 10.0\% \\
\bottomrule
\end{tabular}
\end{table}

Table~\ref{tab:pattern_distribution} should be read as a descriptive allocation of dominant design bundles, not as evidence that the corpus decomposes into sharply separated latent classes. Its role is to indicate which centers of gravity are most common once the project-level coding matrix is re-read at the level of bundled commitments rather than single decisions.

The five patterns below were retained because each satisfied the criteria in Table~\ref{tab:pattern_retention_criteria} and captured a meaningfully different combination of scope, operational commitment, and intended audience. Projects may move from one pattern toward another as their requirements evolve, and borderline cases were resolved by asking which bundle best explained the project's dominant infrastructural commitments rather than every local feature. Accordingly, the representative projects named in the following subsections should be read as dominant assignments rather than as pure or exclusive embodiments of one pattern. Analytically, what matters is that these five patterns make visible the major architectural pathways observed in the corpus.

\subsection{Pattern 1: Lightweight Tool Pattern}
\textbf{Pattern 1: Lightweight Tool}

\textbf{Complexity Level:} Low. These projects usually remain small in scope and operationally simple.

\textbf{Typical Positioning:} Personal tools, quick prototypes, single-purpose assistants, and educational systems.

\textbf{Core Design Bundle:}
\begin{itemize}
    \item No subagent support (single-agent architecture only)
    \item In-memory or simple file-based persistence
    \item Hard-coded or decorator-based tool registration
    \item No sandboxing or basic command filtering only
    \item Imperative control flow
\end{itemize}

\textbf{Example Projects:} subzeroclaw, shrew, and babyclaw illustrate this pattern's emphasis on narrow scope, limited infrastructure, and low operational overhead.

\textbf{Primary Tradeoff:} Minimal architectural overhead comes at the cost of weak support for long-running coordination, durable state, and operational control.

\textbf{Why These Choices Fit Together:} When the workflow is narrow and the project boundary is clear, designers can avoid the overhead of delegation hierarchies, durable state management, and elaborate execution controls. The resulting architecture privileges simplicity, fast iteration, and ease of deployment.

\textbf{Boundary of Applicability:} This pattern works well for narrowly scoped tasks, but it becomes strained when workflows grow longer, tool surfaces expand, or task decomposition becomes important.

\subsection{Pattern 2: Balanced CLI Framework Pattern}
\textbf{Pattern 2: Balanced CLI Framework}

\textbf{Complexity Level:} Low to medium. These projects go beyond single-purpose tools but usually stop short of full multi-layer platforms.

\textbf{Typical Positioning:} Developer-facing command-line assistants, coding tools, and extensible productivity frameworks.

\textbf{Core Design Bundle:}
\begin{itemize}
    \item Basic or tool-based delegation subagent architecture
    \item File-based context persistence (JSONL, Markdown)
    \item MCP-first or decorator-based tool registration
    \item Process-level sandboxing with command filtering
    \item Declarative configuration for tools
    \item Category-based or semantic tool routing
\end{itemize}

\textbf{Example Projects:} openclaw, fast-agent, and kimi-cli are representative of this pattern's emphasis on developer tooling, extensibility, and operational practicality.

\textbf{Primary Tradeoff:} The pattern gains practical extensibility and repeatability, but stops short of the deeper coordination and governance required by larger platforms.

\textbf{Why These Choices Fit Together:} This pattern balances extensibility with deployability. It adopts enough persistence, tooling structure, and safety control to support repeated real use, while avoiding the heavier coordination and governance mechanisms seen in larger orchestrator or enterprise systems.

\textbf{Boundary of Applicability:} The pattern is well suited to developer tooling and medium-complexity workflows, but it may become insufficient when the framework needs deep subagent hierarchies, richer policy control, or organization-wide governance.

\subsection{Pattern 3: Multi-Agent Orchestrator Pattern}
\textbf{Pattern 3: Multi-Agent Orchestrator}

\textbf{Complexity Level:} Medium to high. These projects explicitly invest in task decomposition, role separation, and coordination infrastructure.

\textbf{Typical Positioning:} Complex automation frameworks, collaborative coding systems, research workflow engines, and multi-step execution platforms.

\textbf{Core Design Bundle:}
\begin{itemize}
    \item Orchestrator-worker or recursive subagent architecture
    \item Hierarchical or hybrid memory systems
    \item Structured tool delegation with category routing
    \item Policy-based security with approval workflows
    \item Container-based or WASM sandboxing
    \item Event-driven or structured workflow orchestration
\end{itemize}

\textbf{Example Projects:} claude-code-src, docker-agent, and agentpool exemplify frameworks where task decomposition and coordination become first-class architectural concerns.

\textbf{Primary Tradeoff:} The pattern expands problem-solving reach through coordination, but it also raises the cost of context management, safety design, and system comprehensibility.

\textbf{Why These Choices Fit Together:} Once a project treats coordination as a first-class problem, other dimensions tend to change with it. Context management becomes more explicit, tool use becomes more structured, and safety controls become harder to treat as optional because more components participate in execution.

\textbf{Boundary of Applicability:} This pattern is appropriate when decomposition and coordination create real benefits. For smaller or highly repetitive workflows, its added architectural weight may not be justified.

\subsection{Pattern 4: Enterprise Full-Featured Pattern}
\textbf{Pattern 4: Enterprise Full-Featured}

\textbf{Complexity Level:} High. These projects occupy the most operationally demanding region of the design space.

\textbf{Typical Positioning:} Production deployments, security-sensitive systems, enterprise integrations, and compliance-oriented platforms.

\textbf{Core Design Bundle:}
\begin{itemize}
    \item Multi-level recursive or event-driven subagent architecture
    \item Enterprise-grade memory with vector database integration
    \item Full MCP ecosystem integration
    \item Multi-layer defense: container isolation, policy engines, audit trails
    \item Hierarchical memory with working, short-term, and long-term layers
    \item Plugin architecture with versioning and dependency management
\end{itemize}

\textbf{Example Projects:} openhands, openfang, and nullclaw illustrate this pattern's emphasis on layered controls, extensibility, and operational governance.

\textbf{Primary Tradeoff:} High assurance and extensibility are achieved through substantially greater infrastructure, governance, and organizational cost.

\textbf{Why These Choices Fit Together:} In enterprise settings, harnesses must support not only capabilities but also governance. As a result, extensibility, persistent context, security review, isolation, and auditability appear together as a coherent package rather than as independent add-ons.

\textbf{Boundary of Applicability:} This pattern fits organizations that can absorb its infrastructure and governance cost. For small teams or lightweight products, the same design bundle may be unnecessarily heavy.

\subsection{Pattern 5: Scenario-Verticalized or Research-Oriented Pattern}
\textbf{Pattern 5: Scenario-Verticalized or Research-Oriented}

\textbf{Complexity Level:} Variable. These projects may be simple in infrastructure but sophisticated in the dimension most relevant to their target scenario.

\textbf{Typical Positioning:} Academic research, algorithm prototyping, specialized workflow systems, and vertical tools optimized for a narrow domain.

\textbf{Core Design Bundle:}
\begin{itemize}
    \item Highly variable subagent architectures
    \item Highly variable memory systems
    \item Often minimal or experimental tool implementations
    \item Security often minimal or absent
    \item Sophisticated core capabilities matching research focus
    \item Simplified infrastructure and deployment
\end{itemize}

\textbf{Example Projects:} autoresearchclaw, deer-flow, and deepagents illustrate how this pattern favors experimentation, scenario-specific depth, or methodological exploration over infrastructure completeness.

\textbf{Primary Tradeoff:} The pattern permits rapid experimentation or narrow optimization, but often sacrifices generality, operational robustness, or governance completeness.

\textbf{Why These Choices Fit Together:} These projects optimize for a local objective such as experimentation speed, a domain-specific workflow, or a novel coordination mechanism. As a result, one dimension may be highly developed while other dimensions remain intentionally minimal. Unlike the multi-agent orchestrator pattern, the dominant assignment here is driven less by broad coordination infrastructure and more by specialization around a narrow research or vertical objective.

\textbf{Boundary of Applicability:} The pattern is effective when the project goal is exploration or vertical specialization. It becomes less suitable when the same system must generalize across use cases or support strong operational guarantees.


\section{Discussion}
\label{sec:discussion}

\subsection{Implications for Designers and Selectors}
The design space suggests more than a catalog of options. It also yields a practical decision framework for framework builders and selectors by making visible which commitments tend to travel together and which kinds of complexity remain optional. The points below combine descriptive findings from the corpus with design guidance derived from those findings; they are not intended as universal prescriptions. Their rationale comes from the combination of Sections~\ref{sec:designspace}, \ref{sec:correlations}, and \ref{sec:patterns}: once recurring bundles and recurring tradeoffs are visible across the corpus, they can be used as a decision aid for future architectural choices.

\textbf{For designers, the first implication is to define the intended complexity envelope early.} A lightweight personal tool, a developer-facing CLI, a research workflow engine, and an enterprise platform do not require the same architectural commitments. In the corpus, these differing targets are associated with meaningfully different bundles of decisions, as shown by the separation between recurrent pattern centers in Table~\ref{tab:pattern_distribution}. Designers should therefore begin by clarifying whether the target system is intended for single-step assistance, long-running workflows, multi-agent coordination, or production governance. The key point is that architectural complexity is usually easier to manage when it is chosen as part of an overall operating model rather than accumulated opportunistically feature by feature, because the corpus shows that different positioning choices repeatedly align with different bundled commitments rather than one universal architecture.

\textbf{The second implication is to choose coherent bundles rather than isolated features.} The study shows that decisions about subagents, context handling, tool registration, orchestration, and safety recur in recognizable combinations. In particular, deeper task decomposition is frequently observed alongside more explicit context persistence, and open-ended tool extension is often observed alongside clearer execution boundaries and approval logic; these relationships are summarized in Table~\ref{tab:key_correlations_summary} and elaborated in Sections~5.2 and~5.4. In this setting, architectural coherence matters more than maximizing the number of advanced features, because many seemingly local enhancements are empirically associated with broader coordination and governance burdens elsewhere in the system.

\textbf{For selectors, the main implication is to evaluate fit rather than abstract sophistication.} The patterns in Section~\ref{sec:patterns} indicate that a framework can be well designed for one use case while remaining unsuitable for another, and Table~\ref{tab:pattern_distribution} shows that the corpus does not converge on a single dominant architectural recipe. Selection should therefore consider whether a framework's coordination depth, persistence model, extensibility approach, and safety posture match the operational demands of the target setting. The central evaluative question is not whether a harness is globally ``more advanced,'' but whether its bundle of commitments matches the user's workflow, governance tolerance, and extensibility horizon.

\textbf{A final implication is to treat safety and operability as first-order design concerns.} In the corpus, stronger execution power and broader workspace access are rarely sustainable without at least some corresponding investment in approval logic, isolation, or audit visibility. This is directly visible in the co-occurrence between execution isolation and structured safety control in Section~5.2, and it is reinforced by the negative result in Section~5.3 that capability growth does not automatically produce safety maturity. Sandbox boundaries, approvals, and audit visibility should therefore be designed alongside tool execution and workspace access. For many practical deployments, the decisive question is not what a harness can do in principle, but what it can do under acceptable operational constraints.

\subsection{Relation to Prior Work and Broader Implications}
The findings also clarify how this study relates to prior work and why a design-space perspective adds analytical value. More specifically, they show that Agent harnesses can now be discussed as an architectural domain with recurring internal structure rather than merely as an assortment of rapidly changing framework examples.

\textbf{First, the paper complements capability-oriented agent research.} Much of the Agent literature emphasizes reasoning strategies, planning quality, benchmark performance, or task success. Those contributions remain essential, but they do not by themselves explain how frameworks package tool execution, persistence, delegation, and safety into reusable infrastructure. Our results therefore shift the unit of analysis from agent behavior to the architectural substrate that makes such behavior operationally possible. This shift matters because benchmark-visible behavior can remain similar even when the underlying execution substrate embodies very different assumptions about control, extensibility, and governance.

\textbf{Second, the paper extends existing framework discussions by moving from exemplars to comparative structure.} Prior descriptions of harnesses and frameworks often focus on individual systems or small groups of prominent examples. The present study instead asks which dimensions recur across a broader corpus and which combinations of decisions repeatedly co-occur or fail to co-occur. This shift from case description to structured comparison is what makes a cumulative empirical account possible. It also turns architectural diversity from anecdotal observation into analyzable empirical structure.

\textbf{Third, the results contribute to software-architecture research centered on design decisions.} By treating Agent harnesses as a domain in which infrastructural decisions can be observed comparatively, the paper connects a fast-moving AI-software ecosystem to the longer-standing architectural-decision tradition. In that sense, the contribution is not only descriptive. It also shows that Agent harnesses are mature enough to support cumulative discussion in terms of recurring decisions, bundles, and tradeoffs rather than isolated framework narratives. That is a non-trivial result: it suggests that this part of the Agent ecosystem has reached the point where cross-project architectural regularities can be studied directly.

\textbf{Fourth, the findings support a small set of empirically grounded analytic conjectures.} The first is that higher coordination complexity is frequently observed alongside more explicit context-management services, plausibly because decomposition increases pressure for state transfer, persistence, and summarization. The second is that stronger execution power is frequently observed alongside more explicit governance mechanisms, plausibly because broader action capability raises the operational salience of approval, isolation, and auditability. The third is that more formal tool-registration boundaries are frequently observed in projects pursuing platform-like extensibility or broader ecosystem positioning. These are not causal laws identified by the present study. Rather, they are descriptively grounded conjectures that later work can test longitudinally, experimentally, or across expanded corpora. Their importance lies in converting a descriptive empirical account into a more explicit explanatory agenda without overstating what the current evidence can establish.

\textbf{Finally, the study highlights several recurrent architectural tensions.} The corpus repeatedly reveals tradeoffs between rapid extensibility and operational control, between deeper coordination and manageable context, and between broad execution capability and a credible safety posture. These tensions help explain why the ecosystem does not converge on a single ``best'' architecture and why framework diversity is likely to persist. The broader implication is that pluralism in Agent harness architecture is not merely a transient sign of immaturity; it is at least partly a consequence of genuinely competing infrastructural commitments.

\subsection{Threats to Validity and Limitations}
The main threats to validity follow from the scope and coding strategy of the study.

\textbf{Corpus validity.} Our analysis focuses primarily on open-source projects, with a small number of publicly documented comparison cases, and therefore favors systems that expose enough source-visible structure for architectural comparison. Fully commercial systems, internal enterprise harnesses, and unpublished research prototypes may occupy different parts of the design space. The corpus is therefore broad, but not exhaustive. It was assembled through a systematic but non-PRISMA search and screening process, so the study claims analytic coverage rather than census completeness.

\textbf{Construct validity.} We code observable design decisions from source artifacts and project structure, which improves transparency but does not eliminate interpretive judgment. Some decisions are explicit in code, whereas others must be inferred from coordination logic, persistence pathways, or sandbox boundaries. Borderline cases therefore remain possible. Likewise, the five focal dimensions should be read as analytically stable concerns for this corpus rather than as an exhaustive ontology of Agent-systems architecture.

\textbf{Internal validity.} The co-occurrence analysis is descriptive and does not establish causal direction. A relationship between project scale and architectural complexity, for example, could reflect deliberate planning, emergent growth, team capability, or other confounding factors not directly measured in this study. Coding consistency was strengthened through a sampled secondary review, but chance-corrected agreement was not reported for the full corpus-wide matrix because the study combines mixed variable types and dimension-specific coding structures. The retained archive, however, provides a practical basis for expanding agreement analysis in later extensions of the work.

\textbf{Temporal validity.} The Agent harness ecosystem is evolving quickly. Protocol adoption, orchestration styles, and safety practices may shift substantially over short periods. The present results should therefore be read as a snapshot of a rapidly moving design space rather than as a permanent equilibrium.

\subsection{Future Research Directions}
Several research directions follow directly from the present results.

\textbf{A first direction is longitudinal analysis.} The current paper captures a cross-sectional snapshot, but many of the most interesting questions concern movement through the design space. Future work could track how individual frameworks evolve from lightweight tools into richer platforms and which architectural bundles tend to appear first during that evolution.

\textbf{A second direction is fuller project-level measurement.} The present comparative framework can be extended with more systematic project-level coding across all dimensions, enabling stronger quantitative comparison, broader co-occurrence analysis, and more explicit validation of the patterns identified here.

\textbf{A third direction is comparative evaluation of architectural outcomes.} Once design dimensions are coded more completely, future studies can ask whether certain bundles are associated with maintainability, extensibility, safety incidents, ecosystem growth, or user adoption. That step would move the literature from descriptive empirical characterization toward explanatory and evaluative analysis.

\textbf{A fourth direction is methodological standardization.} The field would benefit from clearer shared codebooks and reusable extraction procedures for harness-architecture analysis. Where appropriate, audit-friendly artifacts can also be exchanged under suitable confidentiality and licensing constraints. Such methodological standardization would make it easier to compare new frameworks against prior work and to turn design-space analysis into a cumulative research program.


\section{Conclusion}
\label{sec:conclusion}

We conducted a protocol-guided, source-grounded empirical study of 70 publicly available agent-system projects in order to characterize the design space of their surrounding non-LLM engineering layer. Rather than reducing frameworks to a single aggregate comparison, the paper asked three questions: which design-decision dimensions recur, what co-occurrences and non-co-occurrences can be observed among those decisions, and how projects with different complexity profiles and product positioning assemble them into typical architectural patterns. Throughout the paper, ``Agent harness'' is used only as a convenient shorthand for this layer rather than as a strict category definition.

\subsection{Research Summary}

The study shows that this surrounding engineering layer can be described through a recurring set of focal dimensions centered on subagent architecture, context management, tool systems, safety mechanisms, and orchestration. These dimensions are not independent. Certain combinations appear repeatedly, such as richer subagent coordination accompanied by more explicit context handling or stronger execution environments paired with more structured safety controls. At the same time, the corpus indicates that different categories of projects occupy different but internally coherent regions of the design space rather than converging on a single architectural form. The resulting picture is therefore neither a ranking nor a maturity ladder, but a structured account of recurring architectural alternatives and the pressures that help hold them together.

\subsection{Contribution Recap}

The paper makes four main contributions. First, it identifies the focal architectural decisions that recur across publicly available Agent execution infrastructures and grounds that characterization in a transparent protocol-guided investigation procedure. Second, it shows that the resulting design space is structured by recognizable co-occurrences and non-co-occurrences rather than by isolated feature selections. Third, it synthesizes typical architectural patterns and uses the empirical findings to formulate analytically useful conjectures about how coordination depth, governance burden, and extensibility ambition shape recurring design bundles. Fourth, it provides an auditable comparative research package through an explicit corpus boundary, search and screening protocol, and operational codebook. Together, these contributions provide an empirical characterization of the surrounding engineering organization of Agent systems rather than a loose collection of implementation anecdotes.

\subsection{Future Work}

Future work can extend this study in several directions. Longitudinal analysis could track how individual frameworks move through the design space over time. Project-level measurement can be refined so that more systematic co-occurrence analysis becomes possible across all dimensions. The analytic conjectures advanced here can also be tested against larger corpora, later snapshots, or more focused subsets of the ecosystem. The present comparative framework may also serve as a foundation for more explicit decision support for framework builders, more systematic comparison of new harness architectures, and eventually more principled architectural benchmarks for Agent infrastructure.

More broadly, this study suggests that the surrounding engineering organization of Agent systems should not be understood merely as incidental implementation wrapping around language models. It exhibits recurring architectural decisions, tradeoffs, and evolution paths across projects that call themselves systems, frameworks, platforms, CLIs, or products. Empirically characterizing that architectural structure is therefore not just descriptive bookkeeping. It is a necessary step toward a more cumulative, comparative, and rigorous study of Agent-systems engineering.



\clearpage
\appendix
\section{Project Corpus}
\label{sec:appendix_projects}

\subsection{Project Selection and Data Collection}

We selected projects according to the following criteria: (1) publicly available implementation evidence, typically source code on platforms such as GitHub, sufficiently detailed public technical documentation, publicly accessible papers, or other source-visible materials already circulating publicly at the time of analysis; (2) meaningful Agent implementation providing core surrounding infrastructure beyond a mere API wrapper or prompt template; (3) active development or significant recent updates within the past 12 months; and (4) sufficient scale for architectural analysis, operationalized as more than approximately 500 lines of code or an equivalent architectural footprint for public-evidence comparison cases.

\subsection{Project Metadata Summary}

\begin{table}[htbp]
\centering
\caption{Project Corpus Overview (N=70)}
\label{tab:project_corpus_summary}
\begin{tabular}{lcc}
\toprule
\textbf{Category} & \textbf{Count} & \textbf{Percentage} \\
\midrule
Total Projects & 70 & 100\% \\
Open-source Repositories & 67 & 95.7\% \\
Publicly Documented or Source-visible but not Fully Open-source & 3 & 4.3\% \\
Official or First-party Products & 6 & 8.6\% \\
Community-led Projects & 64 & 91.4\% \\
\bottomrule
\end{tabular}
\end{table}

\subsection{Complete Project List with Public Evidence Links}

\small
\begin{longtable}{p{2.4cm} p{4.9cm} p{0.9cm}}
\caption{Complete Project List with Public Evidence Links} \label{tab:project_corpus} \\
\toprule
\textbf{Project} & \textbf{Public evidence} & \textbf{Official} \\
\midrule
\endfirsthead
\multicolumn{3}{c}{\tablename\ \thetable\ -- Continued} \\
\toprule
\textbf{Project} & \textbf{Public evidence} & \textbf{Official} \\
\midrule
\endhead
\midrule
\multicolumn{3}{r}{Continued on next page} \\
\bottomrule
\endlastfoot
MASFactory & \url{github.com/BUPT-GAMMA/MASFactory} & N \\
agentpool & \url{github.com/phil65/agentpool} & N \\
angel-claw & \url{github.com/Abdur-rahmaanJ/angel-claw} & N \\
AstrBot & \url{github.com/AstrBotDevs/AstrBot} & N \\
atombot & \url{github.com/daegwang/atombot} & N \\
auto-dev & \url{github.com/phodal/auto-dev} & N \\
autobot & \url{github.com/charliermarsh/autobot} & N \\
AutoResearchClaw & \url{github.com/aiming-lab/AutoResearchClaw} & N \\
babyclaw & \url{github.com/yogesharc/babyclaw} & N \\
claude-code-src & No official public repository; source-visible leaked snapshot & Y \\
claw-code & \url{github.com/ultraworkers/claw-code} & N \\
claw0 & \url{github.com/shareAI-lab/claw0} & N \\
clawdroid & \url{github.com/clawdroidxyz/clawdroid} & N \\
clawlet & \url{github.com/mosaxiv/clawlet} & N \\
cline & \url{github.com/cline/cline} & N \\
code-assistant & \url{github.com/stippi/code-assistant} & N \\
codex & \url{github.com/openai/codex} & Y \\
CoPaw & \url{github.com/agentscope-ai/CoPaw} & N \\
deepagents & \url{github.com/langchain-ai/deepagents} & N \\
deer-flow & \url{github.com/bytedance/deer-flow} & N \\
docker-agent & \url{github.com/docker/docker-agent} & N \\
droidclaw & \url{github.com/unitedbyai/droidclaw} & N \\
everything-claude-code & \url{github.com/affaan-m/everything-claude-code} & N \\
EvoScientist & \url{github.com/EvoScientist/EvoScientist} & N \\
fast-agent & \url{github.com/evalstate/fast-agent} & N \\
flowlyai & \url{github.com/Nocetic/flowly} & N \\
fount & \url{github.com/steve02081504/fount} & N \\
gemini-cli & \url{github.com/google-gemini/gemini-cli} & Y \\
goose & \url{github.com/aaif-goose/goose} & N \\
hermes-agent & \url{github.com/NousResearch/hermes-agent} & N \\
hermitclaw & \url{github.com/brendanhogan/hermitclaw} & N \\
HiClaw & \url{github.com/agentscope-ai/HiClaw} & N \\
ironclaw & \url{github.com/nearai/ironclaw} & N \\
kilocode & \url{github.com/Kilo-Org/kilocode} & N \\
kimi-cli & \url{github.com/MoonshotAI/kimi-cli} & Y \\
langroid & \url{github.com/langroid/langroid} & N \\
learn-claude-code & \url{github.com/shareAI-lab/learn-claude-code} & N \\
lettabot & \url{github.com/letta-ai/lettabot} & N \\
microclaw & \url{github.com/microclaw/microclaw} & N \\
mimiclaw & \url{github.com/memovai/mimiclaw} & N \\
minion-code & \url{github.com/femto/minion-code} & N \\
mistral-vibe & \url{github.com/mistralai/mistral-vibe} & Y \\
moltis & \url{github.com/moltis-org/moltis} & N \\
moxxy & \url{github.com/moxxy-ai/moxxy} & N \\
nanobot & \url{github.com/HKUDS/nanobot} & N \\
nanoclaw & \url{github.com/qwibitai/nanoclaw} & N \\
NemoClaw & \url{github.com/NVIDIA/NemoClaw} & N \\
nullclaw & \url{github.com/nullclaw/nullclaw} & N \\
oh-my-openagent & \url{github.com/code-yeongyu/oh-my-openagent} & N \\
openclaw & \url{github.com/openclaw/openclaw} & N \\
opencode & \url{github.com/anomalyco/opencode} & N \\
opencrabs & \url{github.com/adolfousier/opencrabs} & N \\
openfang & \url{github.com/RightNow-AI/openfang} & N \\
OpenHands & \url{github.com/OpenHands/OpenHands} & N \\
pi-mono & \url{github.com/badlogic/pi-mono} & N \\
pickle-bot & \url{github.com/czl9707/pickle-bot} & N \\
picobot & \url{github.com/caravelahc/pico-bot} & N \\
picoclaw & \url{github.com/picoclaw/picoclaw} & N \\
PraisonAI & \url{github.com/MervinPraison/PraisonAI} & N \\
qwen-code & \url{github.com/QwenLM/qwen-code} & Y \\
safeclaw & \url{github.com/princezuda/safeclaw} & N \\
shrew & \url{github.com/Masmedeam/shrew} & N \\
subzeroclaw & \url{github.com/jmlago/subzeroclaw} & N \\
supaclaw & \url{github.com/vincenzodomina/supaclaw} & N \\
trinity-claw & \url{github.com/TrinityClaw/trinity-claw} & N \\
VTCode & \url{github.com/vinhnx/VTCode} & N \\
Yuxi & \url{github.com/xerrors/Yuxi} & N \\
zclaw & \url{github.com/tnm/zclaw} & N \\
zeptoclaw & \url{github.com/qhkm/zeptoclaw} & N \\
zeroclaw & \url{github.com/zeroclaw-labs/zeroclaw} & N \\
\bottomrule
\end{longtable}
\normalsize

\subsection{Official Products from Major AI Labs}

Six projects (8.6\%) are official or first-party products from major AI companies:

\begin{itemize}
    \item \textbf{claude-code-src} (Anthropic) -- closed-source product; analyzed via public product materials together with a source-visible leaked snapshot rather than an official public repository
    \item \textbf{codex} (OpenAI) -- \url{github.com/openai/codex}
    \item \textbf{Gemini CLI} (Google) -- \url{github.com/google-gemini/gemini-cli} (Apache 2.0)
    \item \textbf{Kimi CLI} (Moonshot AI) -- \url{github.com/MoonshotAI/kimi-cli}
    \item \textbf{Qwen Code} (Alibaba) -- \url{github.com/QwenLM/qwen-code}
    \item \textbf{Mistral Vibe} (Mistral AI) -- \url{github.com/mistralai/mistral-vibe}
\end{itemize}

\subsection{Notes on Project Names}

Some projects in our corpus have similar names but different origins:
\begin{itemize}
    \item \textbf{claude-code-*} variants: Anthropic's Claude Code is a closed-source product. The corpus uses \texttt{claude-code-src} as a label for the source-visible leaked snapshot analyzed alongside public product materials; other entries (\texttt{everything-claude-code}, \texttt{learn-claude-code}) are community projects.
    \item \textbf{claw-*} variants (claw0, clawdroid, clawlet, hiclaw, ironclaw, microclaw, nanoclaw, nemoclaw, nullclaw, picoclaw, subzeroclaw, supaclaw, trinity-claw, zclaw, zeptoclaw, zeroclaw): These are community variants or related projects to the OpenClaw ecosystem.
    \item \textbf{-bot} variants (autobot, atombot, nanobot, picobot, pickle-bot): These are separate projects with similar naming patterns but distinct origins.
\end{itemize}

\section{Operational Codebook}
\label{sec:appendix_codebook}

This appendix summarizes the operational coding categories used to transform source-visible implementation evidence into the comparative records discussed in Sections~\ref{sec:methodology}--\ref{sec:patterns}. The purpose is not to claim an exhaustive ontology of Agent-systems architecture, but to make the manuscript's focal variables more auditable and easier to extend in later replications.

\subsection{Coding Principles}

Each project record was created from source-first inspection. When possible, variables were coded directly from implementation artifacts such as repository structure, tool-registration code, workflow definitions, persistence layers, and safety boundaries. When the boundary between categories was ambiguous, we attached a confidence note rather than forcing a high-certainty label.

\subsection{Focal Variables and Operational Categories}

\small
\begin{longtable}{p{3.0cm}p{9.0cm}}
\caption{Operational Codebook for the Five Focal Dimensions}
\label{tab:operational_codebook} \\
\toprule
\textbf{Dimension} & \textbf{Operational categories used in coding} \\
\midrule
\endfirsthead
\multicolumn{2}{c}{\tablename\ \thetable\ -- Continued} \\
\toprule
\textbf{Dimension} & \textbf{Operational categories used in coding} \\
\midrule
\endhead
\midrule
\multicolumn{2}{r}{Continued on next page} \\
\bottomrule
\endlastfoot
Subagent architecture &
Presence or absence of subagents; dominant structural pattern coded as None, Basic Spawn, Tool-based Delegation, Pipeline/Stage, Orchestrator-Worker, Multi-level Recursive, Swarm/Collective, or Event-driven. Additional notes recorded creation style, nesting depth, and communication boundary. \\

Context management &
Dominant context-management bundle coded as Context Window, LLM Summarization, File Persistence, Vector Database/RAG, Hierarchical, Hybrid, or Enterprise. Additional notes recorded persistence scope, compression strategy, and token-awareness behavior when these were visible in implementation artifacts. \\

Tool system &
Dominant tool-system pattern coded as Minimalist, Registry, Decorator-driven, Declarative/DSL, MCP-first, Plugin Ecosystem, Enterprise, or Delegation/Proxy. Additional notes recorded how tools were surfaced for discovery and where execution boundaries were enforced. \\

Safety mechanisms &
Approval logic coded descriptively as absent, confirmation-oriented, or policy-structured when visible. Isolation level coded as No isolation, Process separation, Container isolation, or WASM sandboxing. Audit capability coded as No audit, Basic logging, Structured audit, or Tamper-evident logging. \\

Orchestration &
Workflow definition coded as Imperative, Declarative/YAML/DSL, or Event-driven. Planning approach coded as ReAct-style, Plan-and-Execute, or Hierarchical. Additional notes recorded whether orchestration remained local to one loop or coordinated multiple roles, stages, or triggers. \\
\end{longtable}
\normalsize

\subsection{Evidence Trail}

For each project, the internal record retained the coded values, the implementation artifacts supporting those values, and a short note when evidence was partial or interpretive. Appendix~\ref{sec:appendix_projects} provides the full corpus list so that later work can revisit individual cases with the same coding frame.

\section{Search and Selection Protocol}
\label{sec:appendix_search_protocol}

This appendix makes the corpus-discovery procedure more explicit. The goal of the study was broad analytic coverage of a rapidly evolving implementation ecosystem rather than PRISMA-style reconstruction of a closed literature universe. Even so, project discovery followed a repeatable protocol rather than ad hoc collection.

\subsection{Discovery Channels}

Candidate projects were assembled from four channels:
\begin{enumerate}
    \item public repository search, primarily on GitHub and comparable public code-hosting surfaces;
    \item references encountered during related-work review and framework documentation reading;
    \item manual review of widely discussed Agent frameworks, coding agents, and orchestration tools; and
    \item snowball expansion from already retained projects to adjacent frameworks, forks, competitors, and explicitly referenced alternatives.
\end{enumerate}

\subsection{Keyword Families}

Repository search and manual discovery repeatedly used keyword families such as the following:
\begin{itemize}
    \item ``AI agent''
    \item ``agent''
    \item ``harness''
    \item ``vibe coding''
    \item ``agent framework''
    \item ``agent platform''
    \item ``agent system''
    \item ``coding agent''
\end{itemize}

Keyword search was supplemented by project-name tracing when a candidate framework repeatedly appeared in framework comparisons, technical blog posts, benchmark discussions, or repository-level references from already retained projects.

\subsection{Starting Points for Snowball Expansion}

Snowball expansion began from widely discussed frameworks and execution substrates identified during the initial search and related-work pass. Representative early anchors included LangChain, AutoGPT, CrewAI, OpenHands, OpenClaw, Claude Code CLI, and docker-agent. These projects were not treated as a canonical benchmark set; they were used only as practical entry points for discovering adjacent implementations, forks, and alternatives visible through repository references and ecosystem discussion.

\subsection{Screening Logic}

Projects were retained only when they satisfied all of the following conditions:
\begin{enumerate}
    \item publicly available implementation evidence, detailed public technical documentation, or other source-visible material sufficient for architectural inspection;
    \item meaningful execution-substrate functionality such as tool mediation, orchestration, persistence, workspace interaction, reusable delegation, or governance controls;
    \item implementation scale usually exceeding approximately 500 lines of code, or an equivalent architectural footprint for public comparison cases; and
    \item enough observable structure to support comparative coding under the paper's five focal dimensions.
\end{enumerate}

Projects were excluded when they were primarily prompt collections, thin API wrappers, benchmark-only artifacts, very small demonstrations, or products lacking enough public implementation evidence to support source- or paper-grounded coding.

\subsection{Corpus Freeze and Boundary}

The project list was frozen on 23 March 2026 to provide a stable cross-sectional snapshot. The retained corpus therefore reflects the public state of the ecosystem visible up to that date rather than a continuously updated benchmark roster. Appendix~\ref{sec:appendix_projects} reports the resulting corpus members.

\subsection{Why This Is Not PRISMA}

The present study is an empirical analysis of a rapidly changing implementation ecosystem rather than a formal systematic literature review. For that reason, the protocol privileges cross-project architectural coverage, repeatable screening logic, and auditable project records over exhaustive reconstruction of every possible eligible repository. The tradeoff is explicit: the study aims to characterize recurrent architectural structure in a broad corpus, not to claim a complete census of all Agent frameworks ever released.

\end{document}